\documentclass[journal]{IEEEtran}
\usepackage{amsmath,amsfonts}
\usepackage{amssymb}
\usepackage{algorithmic}
\usepackage{array}
\usepackage[caption=false,font=normalsize,labelfont=sf,textfont=sf]{subfig}
\usepackage{textcomp}
\usepackage{stfloats}
\usepackage{url}
\usepackage{verbatim}
\usepackage{graphicx}
\usepackage{multirow}
\usepackage{threeparttable} 
\usepackage{subfig}
\usepackage{fdsymbol}

\newtheorem{definition}{Definition}
\newtheorem{hypothesis}{Hypothesis}

\def\BibTeX{{\rm B\kern-.05em{\sc i\kern-.025em b}\kern-.08em
    T\kern-.1667em\lower.7ex\hbox{E}\kern-.125emX}}
\usepackage{balance}
\begin{document}
\title{Learning a Structural Causal Model for Intuition Reasoning in Conversation}
\author{Hang Chen, Bingyu Liao, Jing Luo, Wenjing Zhu, ~\IEEEmembership{Xinyu Yang}

\thanks{The authors are with the Department of Computer Science and Technology, 
Xi'an Jiaotong University, Xi'an Shannxi, 710049; Department of Cultural and Religious Studies, Chinese University of Hong Kong, Hong Kong, Shatin 999077. 
and the Du Xiao Man Inc, Beijing, 100089.\\
E-mail: albert2123@stu.xjtu.edu.cn, bingyuliao@link.cuhk.edu.hk, luojingl@stu.xjtu.edu.cn, thinkre7@163.com, yxyphd@mail.xjtu.edu.cn}}
\markboth{Journal of \LaTeX\ Class Files,~Vol.~18, No.~9, May~2023}%
{How to Use the IEEEtran \LaTeX \ Templates}

\maketitle

\begin{abstract}
Reasoning, a crucial aspect of NLP research, has not been 
adequately addressed by prevailing models including Large Language Model. 
Conversation reasoning, as a critical component of it, 
remains largely unexplored due to the absence of 
a well-designed cognitive model. 
In this paper, inspired by intuition theory on 
conversation cognition, we develop a conversation cognitive model (CCM)
that explains how each utterance receives and activates 
channels of information recursively. Besides, 
we algebraically transformed CCM into a structural causal model (SCM) 
under some mild assumptions, rendering it compatible 
with various causal discovery methods. 
We further propose a probabilistic implementation 
of the SCM for utterance-level relation reasoning. 
By leveraging variational inference, it explores 
substitutes for implicit causes, addresses the issue of 
their unobservability, and reconstructs the causal representations 
of utterances through the evidence lower bounds. 
Moreover, we constructed synthetic and simulated datasets 
incorporating implicit causes and complete cause labels, 
alleviating the current situation where all available datasets 
are implicit-causes-agnostic. 
Extensive experiments demonstrate that our proposed method 
significantly outperforms existing methods 
on synthetic, simulated, and real-world datasets. 
Finally, we analyze the performance of CCM 
under latent confounders and propose theoretical ideas 
for addressing this currently unresolved issue.

\end{abstract}


\section{Introduction}

Reasoning tasks have historically exhibited sluggish progress 
in Natural Language Processing (NLP)
~\cite{gupta2019neural,hunter2000reasoning}. 
Despite improving performance across various downstream NLP tasks 
facilitated by recent advancements in large language models (LLMs), 
reasoning tasks continue to present persistent challenges
~\cite{kasneci2023chatgpt,wei2022chain}. 
Consequently, given the current mainstream research, 
we classify reasoning tasks 
into two distinct categories: \textbf{rational reasoning}
~\cite{kojima2022large,zelikman2022star}, 
and \textbf{intuitive reasoning}
~\cite{chen2023affective,zhao2022cauain}.

\textbf{Rational reasoning} is always employed to solve problems 
 relying on mathematical knowledge or fundamental facts, 
such as primary math problems. 
A question example from an arithmetic reasoning dataset
~\cite{roy2016solving} can illustrate this: 

\textbf{Q}: \textit{A chef needs to cook 15 potatoes. 
He has already cooked 8. If each potato takes 9 minutes to cook, 
how long will it take him to cook the rest?} 

As suggested by CoT~\cite{zhang2022automatic}, many researchers
~\cite{wang2022rationale,li2022advance}
adopt the identical causal chain to ``step-by-step'' 
derive intermediate results ($15-8=7$) 
and ultimately arrive at the solution ($9*7=63$). 

However, \textbf{intuitive reasoning} does not rely on widely 
fundamental facts to support reasoning 
but on semantic perception and abstract concepts. One pervalent area 
is conversation reasoning, 
as demonstrated in dialogue generation tasks
~\cite{li2017dailydialog}: 

\textit{speaker 1:``I can't stand the stupid guy any longer. 
It's unbelievable!''}

\textit{speaker 2:``Oh, my dear lady. Take it easy. 
You should forgive a green hand like him.''}

\textbf{Q}: \textit{What should speaker 1 say to respond to 
speaker 2?}

In the realm of arithmetic reasoning, 
the path to the final result is typically unambiguous, 
with clear intermediate steps leading to the ultimate outcome. 
researchers can come to a conclusion of  
``what will be led to by known knowledge" and 
``which information will guide to the answer". 
However, as shown in the case of conversation reasoning, 
considerable discourse exists regarding the factors 
influencing the content of a given utterance: 
\cite{chen2022learning,sordoni2015neural} 
support that previous dialogue content plays a pivotal role in 
determining the current utterance, 
\cite{peng2019topic,zhou2018emotional} argue that 
the emotional dynamics of the speaker 
greatly influences their response, 
and external knowledge such as the speaker's memory or beliefs 
is posited to be essential in shaping the current utterance 
~\cite{chen2023affective,zheng2020difference}. Hence, 
within tasks related to intuitive reasoning (e.g., conversation reasoning), 
the mainstream research on computational models lacks the 
same level of consensus as observed in rational reasoning
~\cite{mahowald2023dissociating}. 

Therefore, in this article, 
based on the aforementioned issues in intuitive reasoning, we focus 
on a specific area of it$-$conversation reasoning, 
and investigate the following questions: 

\begin{itemize}
  \item How to construct a cognitive model that 
  elucidates the diverse existing intuitive theories 
  about dialogue interpretation?
  \item How to algebraize the proposed cognitive model 
  and implement it through a deep learning network?
  \item How to evaluate the effectiveness and interpretability 
  of the model processing conversation data?
\end{itemize}

For the first question, we have developed a 
conversation cognition model (CCM) that draws inspiration from 
a wealth of intuitive theories in psychology~\cite{baker2014modeling}, 
linguistics~\cite{funakoshi2022non}, 
and sociology~\cite{litman1987plan}. 
Among them, common ground~\cite{stalnaker2002common}, as a term 
for the presumed background information shared by participants 
in a conversation, is influenced by the speaker's perception 
of the preceding discourse and their mental state, 
which drives the conversation toward achieving 
mutual understanding and agreement. This shared knowledge, 
as an essential intermediate state in conversation reasoning, 
integrates a plethora of external and internal motivations, thereby 
serving as a unified explanation for utterances and actions 
~\cite{allan2013common,horton1996speakers}. 
Expanding on this foundational intuition theory, 
CCM explains: 1) how speakers assimilate contextual 
perception, as well as information about their beliefs, emotions, 
and other mental states, and externally unify them as their 
expressions of utterances and actions; and 2) how the utterance 
and action generated in a conversation activate 
various channels of information to modify subsequent plans. 
Additionally, 
we employ the probability formulation of CCM to explain the 
inference process underlying several prevalent conversation-related tasks. 

For the second question, to render CCM a computable model, we employed assumptions 
under intermediate variables~\cite{spirtes2013calculation} 
and non-descendant child nodes~\cite{bernstein2020ordering} 
for simplification, subsequently transforming it into a 
structural causal model (SCM). The SCM indicates that 
each utterance is influenced by two factors, 
namely endogenous variables (other utterances) 
and exogenous variables (internal states). 
Our transformation process preserves many aspects of 
alignment between SCM and CCM, such as the observable variables 
in SCM, which represent the concepts in CCM 
that can be mutually comprehended by both the speaker and the observer. 
Given the limitations imposed by existing dialogue resources 
and data collection, we have denoted endogenous variables 
as explicit causes, representing influences in utterances 
being known and observable, and exogenous variables 
as implicit causes, symbolizing the hidden and unobservable 
influence of mental states on utterances. This enables us to 
use Graph Neural Network (GNN)~\cite{scarselli2008graph} 
to compute the influence of the causal strength matrix 
and to obtain a reconstructed representation of causality 
via variational inference. Specifically, 
we consider the implicit causes as latent variables 
and use GNN to compute the product of the input utterance embeddings 
and the SCM coefficient matrix. Then, we use the product of 
the sampled implicit causes and the causal strength matrix 
as the reconstructed utterance representation. 
The adjacency matrices in the encoder and decoder 
align with the adjacency matrix in the 
linear SCM autoregressive form,  thus ensuring interpretability 
during the inference process.

For the third question, we have created a synthetic dataset 
with exogenous variables and a simulated dataset 
with complete causal relationships to alleviate the 
scarcity of evaluative datasets for conversation reasoning. 
The synthetic dataset is constructed through DAG sampling, 
while the simulated dataset is 
generated by GPT-4~\cite{openai2023gpt4} with 
specified causal relationships for the dialogue samples. 
Subsequently, we conducted extensive experiments to evaluate the 
performance of our method on both explicit and latent causes. 
The evaluation encompassed unsupervised large language models 
(GPT-3.5, GPT-4), supervised pre-trained models (RoBERTa), 
and state-of-the-art methods in relevant tasks. 
The results not only reveal the limited performance of 
existing models in conversation reasoning, 
but also highlight the significant enhancements 
brought about by our cognitive model and inference method.

In the discussion, we extensively examined the issue of 
hidden confounders that currently elude resolution 
and put forth some experimental designs to provide 
useful insights for future research.

In summary, our contributions can be summarized as a cognitive model 
and causal explanation for conversation, an algebraical SCM 
transformation and a probabilistic implementation of it, 
evaluation datasets for conversation reasoning, 
and comprehensive experiments.  

\section{Related Work}

\subsection{Intuitive Reasoning with Causal Relationships}
Intuitive reasoning and learning with causal relationship has become 
an active research area since they have shown superior performances 
over causality-agnostic counterparts in recent years
~\cite{9690010,9904301,9051845}. 
One promising approach is to design an empirical or cognitive model 
for the particular text type by encouraging related variables or 
attributes to be participants in causal discovery. 
In online textual advertisements~\cite{9833343}, 
a context-aware persuasion model 
has been proposed by~\cite{9556599}, which probes the causal 
relationship between persuasive tactics and product attributes.  
In the recommendation system~\cite{9681226,9501971}, 
CaDSI~\cite{9736612} causally models 
the underlying relationship of the recommendation system with users' true intents 
and thus learns semantics-aware representations 
without confounders stemming from context information. 
In other modeled text types, more research is involved in 
analyzing how variables outside the model contribute to 
biases or confounding effects on the entire model. Such as knowledge 
graph~\cite{9512424,8974254,8890615,9158381}, User interaction attributes, 
and preference learning are considered as exogenous variables 
and latent confounders of the knowledge structure~\cite{9996555}. 
Furthermore, there are more diverse text types 
that have constructed specific causal models, 
such as event records~\cite{9424449,9963576}, 
short text~\cite{5677516,9996587},
expert systems~\cite{9099622}, 
and intelligent applications~\cite{5416723,8769915}.

Among them, conversation reasoning has only been explored 
in certain specific scenarios, such as 
question-answering conversation~\cite{10102595,9404866,9104878}, 
or emotion-causal reasoning
~\cite{chen2022learning,xia-ding-2019-emotion}. 
However, more general conversation patterns 
have not been fully explored 
due to the lack of a universally computed cognitive model 
~\cite{ghosal-etal-2019-dialoguegcn,shen-etal-2021-directed}
and datasets with complete inference relationships~\cite{poria2021recognizing}. 
Therefore, we have designed a conversation cognitive model that  
hopes to provide useful insights for conversation reasoning research. 
We have also created a simulated dialogue dataset that includes 
all causal relationships to fill the gap in research data.
\subsection{Conversation cognitive}

The study of human communication has attracted the attention of 
scholars from diverse fields such as psychology
~\cite{van1983strategies,sidera2018theory}, 
linguistics~\cite{heim1983file,hunt2022s}, and artificial intelligence
~\cite{poesio1997conversational}.  
Over the years, quite a few models have been put forth by 
researchers to elucidate the intricacies of dialogue
~\cite{kamp2013theory,brown2016memory,chen2022toward}.

While the precise interpretations of these models may differ, 
they share a common thread~\cite{mckinley2017memory}. 
These models collectively highlight 
that conversations encompass far more than a mere succession of 
utterances produced turn by turn in a conversation. 
Conversations rely on common ground, 
including perceptive aspects such as mutual knowledge, 
assumptions~\cite{clark1991grounding}, presuppositions
~\cite{stalnaker1978assertion}, memory~\cite{brown2016memory}, 
and mental states such as mutual beliefs and emotions
~\cite{kleres2011emotions}, 
the expression of which within narratives and conversations 
has garnered significant interest among scholars in psychology 
and sociology throughout the past decades~\cite{hargie2021skilled}. 
Additionally, 
both the presenter and accepter of the same conversation 
contribute content that reflects their respective intentions
~\cite{clark1989contributing}. 
As the conversation unfolds, participants continuously update 
their shared understanding and common ground in a dynamic manner. 
The formation of collective actions within a conversation hinges 
upon accumulating this common ground. 
However, certain complex situations arise that necessitate 
additional measures. For instance, 
when speakers require presuppositions to accept new content, 
a pervasive process known as bridging~\cite{clark1977inferences} 
comes into play, 
enabling the incorporation of supplementary presuppositions. 
Moreover, when participants' beliefs diverge instead of accumulating, 
it becomes imperative for speakers to proactively ensure 
that the content of their assertions contributes positively to 
the common ground~\cite{solomon2012interpersonal}.

Sociologists recognize the significance of investigating 
the advancement of a more intricate theoretical understanding of 
emotions. This entails exploring their nuanced structures of 
meaning, the interplay between overt and covert emotional 
interpretations, and delving into the patterns and progression of 
emotions within discourse~\cite{bloch1996emotions}. 
Nonetheless, it is worth noting that 
these studies heavily rely on empirical data as a foundation 
for their analysis~\cite{solomon2012interpersonal,kleres2011emotions}.

\section{Conversation Cognitive Model} 

In Subsection~\ref{sec3.1}, we introduce the Conversation Cognition Model (CCM) 
and analyze the taxonomy of existing conversational tasks. 
In Subsection~\ref{sec3.2}, we present the Structural Causal Model and 
its corresponding assumptions to transpose the CCM into the 
algebraic-modeling SCM. 
 
\subsection{Model Describing}\label{sec3.1}

\begin{figure*}
  \centering
  \includegraphics[width=0.9\linewidth]{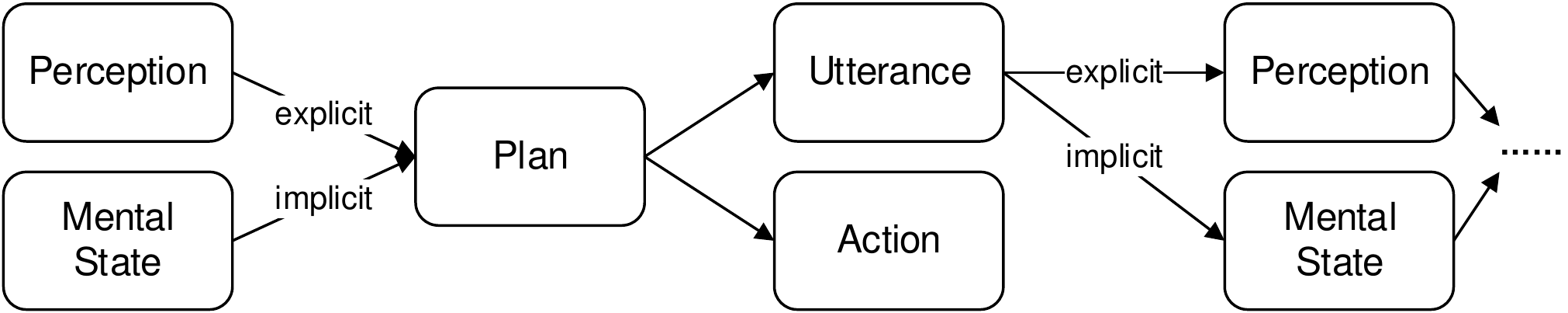}
  \caption{Conversation Cognitive Model (CCM) of an intuitive theory of 
    dialogue. ``Perception'' represents the speaker's understanding of 
    previous utterances. ``Mental State'' represents desires, memory, 
    experience or emotion of the speaker. ``Plan'' stands for the reaction 
    to the latest perception and mental state and is expressed by two 
    external outcomes: ``Utterance'' and ``Action.'' } 
  
  \label{figCCM}
\end{figure*}

Individuals possess abundant intuitive theories to steer the 
observation of others' thinking and behaving, 
enabling them to explicate and deduce underlying motives
~\cite{gopnik1997words}. 
Such intuitive theories consist of a structured ontology 
of concepts (e.g., goals, speech, behavior) and 
the causal relationships relating to these concepts
~\cite{gerstenberg2017intuitive}.
In conversation, individuals (e.g., speaker 1) use their intuitive theory 
of utterances and actions to reason about what others 
(e.g., speaker 2) are going to do 
(e.g., defensive, aggression, ingratiation, and so on.) 
called ``Plan'' in our theory, and thereby determine 
the optimal course of action for 
how best to respond in given social situations. 

Our intuition theory of conversation contains two essential causal 
relationships: the first one connects utterances to their causes, 
the second one involves the outcomes of utterances. According to these 
two relationships, we have developed a Conversation Cognitive Model (CCM) depicted 
in Figure~\ref{figCCM}. In a conversation, the emergence of a plan 
is intuitively triggered by the motivational understanding 
of what speakers talk about (\textbf{Perception}) 
and primarily conveyed through \textbf{Utterance} and \textbf{Action}. 
Apart from \textbf{Perception}, \textbf{Mental State}, 
such as desires, memory, experience, and emotion, 
is also a critical component inducing \textbf{Plan}. 
For simplicity, we omit the influence of the \textbf{Action}. 
As an external interactive expression, the \textbf{Utterance} 
brings about modification to subsequent \textbf{Perception} 
and \textbf{Mental State}. 

Using this CCM, we can explain a taxonomy 
of existing inferences about conversation. 
We use the lowercase initial of each concept in the following 
(e.g., $p$ represents \textbf{Perception}, $pl$ represents \textbf{Plan}). 
Emotion Recognition in Conversation (ERC) task aims to infer the 
probability $P(m|u)$, so most research in this task concerned about 
which utterance contributes the emotion best
~\cite{ghosal-etal-2019-dialoguegcn,Shen2021DialogXLAX,chen2022learning}. 
The plan from different interlocutors $P(m|pl)\propto P(pl|m)P(m)$
is another research point 
~\cite{zhang2019modeling,lian2021decn,shen-etal-2021-directed}. 
Emotion-Cause Pair Extraction (ECPE) task in conversation
~\cite{poria2021recognizing} additionally probes the cause utterance 
of an emotion utterance $P(u|m)P(m)$. Increasing works indicate that 
other related components play critical roles in this reasoning task, 
such as the plan for $P(u|pl,m)P(m)$
~\cite{chen2023affective,gao2021improving} and 
the perception of utterance for $P(u|pl,m,p)P(m)$
~\cite{zhao2022cauain,li2022ecpec}. 
Conversation Generation (CG) task aims to estimate $P(u|pl)$ with 
the appraisals to what agent is going to do. There are many methods 
involving internal influences $P(u|pl,m)$
~\cite{zhou2018commonsense,peng2019topic}, 
external influences $P(u|pl,p)$
~\cite{sun2018emotional,wang2021learning}, or both
~\cite{zhou2018emotional,liu2019knowledge}. In the internal estimation 
of these tasks, $P(m|p,u) \propto P(m)\sum_{pl} P(u|pl)P(pl|p,m)$ is 
a prevalent appraisal.

\subsection{SCM Transformation}\label{sec3.2}

\begin{figure}
  \centering
  \includegraphics[width=0.9\linewidth]{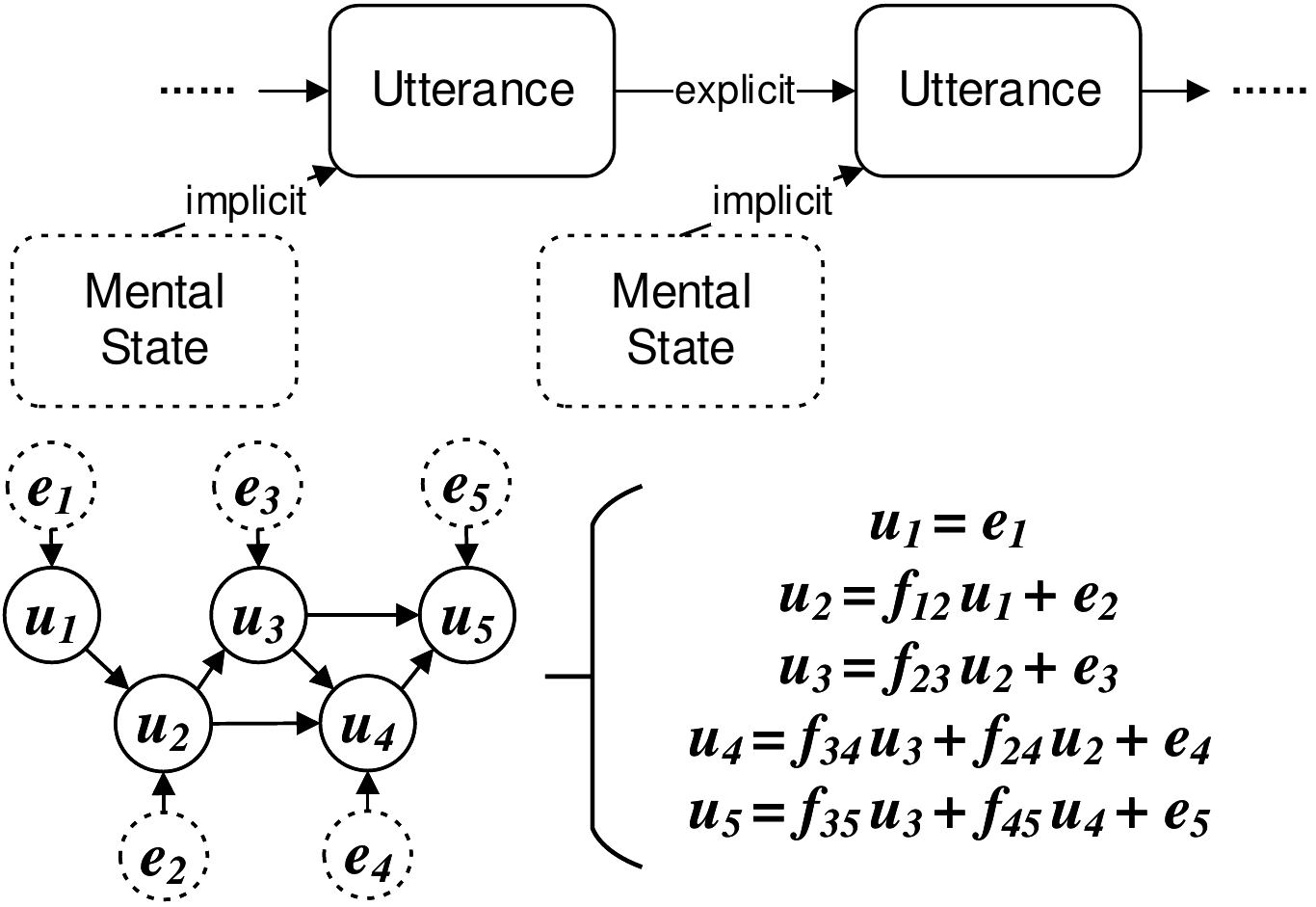}
  \caption{SCM transformed by CCM. 
  The top half of the image shows the cognitive model 
  with causal irrelevant variables omitted, 
  consisting only of explicit utterances and implicit mental states. 
  The bottom half parameterizes this model with a 
  five-utterance conversation example, 
  where $U$ represents utterances, $E$ represents mental states, 
  and subscripts denote the natural sequential index 
  of utterances in the conversation.} 
  \label{figSCM}
\end{figure}
 
In this subsection, we algebraically transform the CCM 
into a Structural Causal Model (SCM), 
which has become a popular choice in the field of causal inference. 
We begin this transformation from the definition of SCM.

\begin{definition}[Structural Causal Model]
  An SCM is a 3-tuple $\langle X, E, \mathcal{F} \rangle $, where 
  $X$ is the entire set of observed variables 
  $X=\{x_{i}\}^{N}_{i=1}$, 
  $E$ is the set of exogenous variables $ E=\{e_{i}\}^{N}_{i=1}$ 
  corresponding to each $x_{i}$, $N$ is the number of variables. 
  Structural equations $\mathcal{F}=\{f_{i}\}^{N}_{i=1}$ are functions that determine 
  $X$ with $x_{i}=f_{i}(Pa(x_{{i}}))+e_{i}$, 
  where $Pa(x_{{i}})\subseteq X$ is the parent set of $x_{i}$. 
  
  \label{def1}
\end{definition}

There are some methods~\cite{shimizu2006linear,sanchez2019estimating} 
built upon the SCMs that overcomes the limitation of 
Markov equivalence classes (e.g., $A \rightarrow B \rightarrow C$ and 
$A\leftarrow B \rightarrow C$). 
The independent conditions enable noise terms and residual terms 
to identify more comprehensive causal relationships 
between the two fitted variables~\cite{chen2023affective}. 
For example, $\Sigma_{X} \Vbar Y, \Sigma_{Y} \nVbar X
\Rightarrow Y \rightarrow X$, while $\Sigma_{X} \nVbar Y, 
\Sigma_{Y} \Vbar X \Rightarrow X \rightarrow Y$ 
($\Sigma$ represents the residual terms in fitting process).

From Definition~\ref{def1}, we would assume that any observed variables 
satisfy the following generating process:

\begin{equation}
  x_{i}=\sum_{x_{j}\in Pa(x_{i})} f_{j,i}x_{j}+e_{i}  
  \label{eqn1}
\end{equation}

SCM defines algebraic relationships between observed variables 
based on an additive noise model. In order to make SCM applicable to 
the CCM, we simplify the CCM by making the following two assumptions.

\begin{hypothesis}[Omitting mediator Nodes]
  If the causal relationship between three nodes 
  $A$, $B$, and $C$ is $A\rightarrow B\rightarrow C$, 
  and $B$ is unobservable and cannot be conditionally controlled, 
  then $B$ can be omitted and $A$, $C$ can be linked as $A \rightarrow C$.
  \label{hyp1}
\end{hypothesis}

\begin{hypothesis}[Omitting Non-descendant Child Nodes]
  If $A$ is a non-descendant node and only has one parent node, 
  then $A$ can be omitted.
  \label{hyp2}
\end{hypothesis}

Hypothesis~\ref{hyp1} simplifies the causal order length 
of the chain model, while Hypothesis~\ref{hyp2} 
merges observed nodes with the same causal properties. 
These hypotheses are broadly used in various studies building SCM 
and have been proposed theoretical proofs
~\cite{xie2020generalized,lowe2022amortized}. 

According to Hypothesis~\ref{hyp2}, we can naturally omit 
\textbf{Action} in CCM, as it is essentially irrelevant 
to our research topic. \textbf{Plan} plays the role 
of a mediator in both chain structures \textbf{Perception} 
$\rightarrow$ \textbf{Plan} $\rightarrow$ \textbf{Utterance} 
and \textbf{Mental State} $\rightarrow$ \textbf{Plan} 
$\rightarrow$ \textbf{Utterance}, and it can only be inferred 
through third-person appraisals (i.e., it is unobservable). 
Therefore, we can also remove \textbf{Plan} and 
build two new connections: \textbf{Perception} 
$\rightarrow$ \textbf{Utterance} and \textbf{Mental State} 
$\rightarrow$ \textbf{Utterance}. 

Although \textbf{Perception} and \textbf{Mental State} are both 
influenced by \textbf{Utterance}, most of the mental state is 
still unobservable. A number of cross-sectional studies 
~\cite{houlihan2022reasoning,teo2022modeling} 
have suggested that observers and speakers can achieve 
significant agreement in \textbf{Perception}. 
However, for \textbf{Mental state}, it is difficult for 
other observers except the speaker to infer it. 
Therefore, we merge the part of \textbf{Mental state} 
that can be effectively third-person estimated 
(such as emotional reactions to the previous utterances) 
with \textbf{Perception} (which is indeed more suitable 
from an observational perspective), 
and ignore this mixture through Hypothesis~\ref{hyp1}. 
As for the unobservable part of \textbf{Mental state}, 
we block the influence of \textbf{Utterance} 
$\rightarrow$ \textbf{Mental state}, making it an exogenous variable. 
This mapping method is called latent projection and is often 
adapted to disentangle unobservable variables
~\cite{agrawal2021decamfounder,evans2016graphs}. 

For example, Speaker 1’s \textbf{Utterance 1}: 
\textit{``I can't stand the 
stupid guy any longer. It's unbelievable!''}
and Speaker 2’s \textbf{Utterance 2}: \textit{``Oh, my dear lady. 
Take it easy. You should forgive a green hand like him.''} 
From these, it's easy to infer that Speaker 2's 
\textbf{Perception} of \textbf{Utterance 1} is anger 
and disappointment. With a \textbf{Mental State} aimed at 
comforting Speaker 1, Speaker 2 \textbf{Plans} to find reasons 
for Speaker 1 to feel relieved. In the SCM, we can understand 
that \textbf{Utterance 2} is both a response to \textbf{Utterance 1} 
and is influenced by the \textbf{Mental State} of providing comfort. 
Both the CCM and SCM reflect the influence of the 
\textbf{Mental State} on \textbf{Utterance 2}, i.e., 
if there was no aim to comfort Speaker 1, 
\textbf{Utterance 2} would more likely have been: \textit{``Yes, 
I completely agree with you, he's really too stupid!''}

All in all, we can simplify CCM to a model with only 
\textbf{Utterance} and \textbf{Mental State}, as shown in Figure
~\ref{figSCM}. \textbf{Utterance} is an observable variable 
that exhibits an explicit relationship with each other, 
while \textbf{Mental State} is an unobservable exogenous variable 
that implicitly affects the corresponding \textbf{Utterance}. 
In the linear setting, each \textbf{Utterance} writes: 

\begin{equation}
  u_{i}=\sum_{j\in rel_{u_{i}}} f_{j,i}u_{j}+e_{i}  
  \label{eqn2}
\end{equation}
where $u_{i}$ represents the $i$-th utterance in a dialogue, 
$e_{i}$ represents its corresponding mental state. 
$rel_{u_{i}}$ denotes a set of utterances that point to the $u_{i}$, 
$f_{j,i}$ stands for the causal strength in 
$u_{j} \rightarrow u_{i}$. Finally, we could give the definition of 
SCM in the conversation. 

\begin{definition}[SCM in the conversation]
  An SCM is a 3-tuple $\langle U, E, \mathcal{F}\rangle $, where 
  $U$ is the set of utterances 
  $U=\{u_{i}\}^{N}_{i=1}$, 
  $E$ is the set of exogenous Mental State $ E=\{e_{i}\}^{N}_{i=1}$ 
  corresponding to each $u_{i}$, $N$ is the number of utterances. 
  Structural equations $\mathcal{F}=\{f_{i}\}^{N}_{i=1}$ are functions that determine 
  $U$ with $u_{i}=f_{i}(rel_(u_{{i}}))+e_{i}$.
  
  \label{def2}
\end{definition}

Definition~\ref{def2} indicates that each utterance 
is explicitly influenced by the previous utterances 
and implicitly influenced by the corresponding mental states. 
Therefore, other utterances and corresponding mental states 
are the causes of the target utterance. Specifically, 
we refer to other utterances as explicit causes 
because it is observable and endogenous in SCM, 
and refer to mental states as implicit causes 
because it is unobservable and exogenous in SCM.

\section{Probabilistic Method} 

\subsection{Variational Inference Framework}

\begin{figure*}
  \centering
  \includegraphics[width=0.9\linewidth]{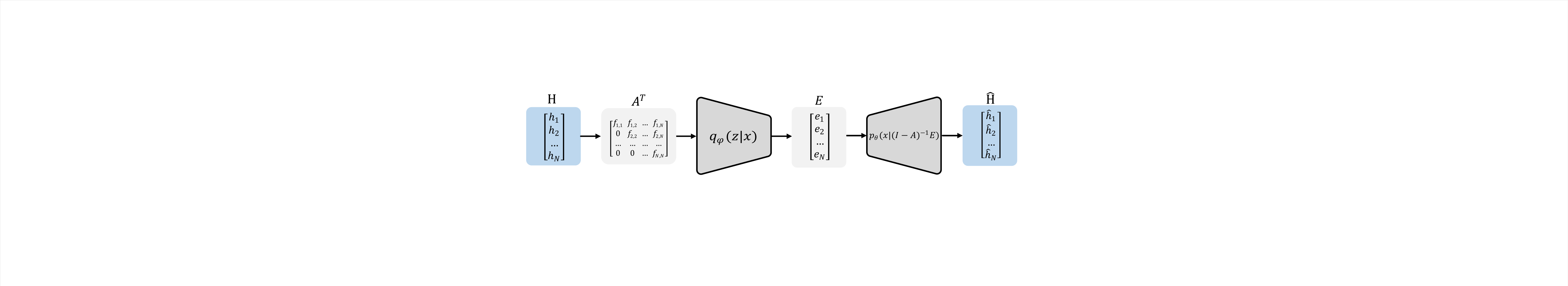}
  \caption{A probabilistic framework of our method. 
  $q_{\varphi}(z|\mathcal{X})$ predicts the implicit causes $E$ from 
  the input $x$ (e.g., exogenous variable matrix in SCM). The decoder 
  $p_{\theta}((x|(I-A)^{-1}E))$ learns to reconstruct $\widehat{X}$ 
  given the $E$ and inverse of predicted $z$. }
  
  \label{figPF}
\end{figure*}

we denote the word embedding of $U$ by $H={h_{1}, h_{2}, \dots, h_{N}}$, 
and the adjacency matrix by $A\in \mathbb{R}^{N \times N}$. 
Equation~\ref{eqn2} describes the generation process of a 
single variable, but it can be trivially represented by 
a matrix form, where each row of $A$ represents the causal strength 
from all variables to one observed variable (e.g., $A^{T}_{i,j}=f_{i,j}$). It is 
a strong inductive bias that $A$ is a strictly lower triangular 
due to the acyclic causal order~\cite{shimizu2006linear}, 
and the relationships between utterances in rows can also be written as : 

\begin{equation}
  H=AH+E 
  \label{eqn3}
\end{equation}
where $E\in \mathbb{R}^{N \times D}$ is the matrix of implicit causes 
$e$. Under this matrix form, generating a random 
independent noise $E$ can be equivalent to ancestral 
sampling from the DAG: $H=(I-A)^{-1}E$. We can build a pair of 
Autoregression Equations: $E=(I-A)H$ and $H=(I-A)^{-1}E$
These two equations describe a general form as a decoder 
of a generation model that takes noise $E$ as input and returns $X$ 
as results and Equation~\ref{eqt4} describes the corresponding 
encoder. Despite the diverse implementation of parameterized 
deep learning networks, we without loss of generality adopt 
$f(\cdot)$ to encapsulate the neural network architecture: 

\begin{equation} 
  E=f_{2}((I-A)f_{1}(H))
  \label{eqt4}
 \end{equation}
 \begin{equation} 
  H=f_{4}((I-A)^{-1}f_{3}(E))
  \label{eqt5}
 \end{equation}
 where $f(\cdot)$ performs nonlinear transforms on $H$ and $E$. 
 Graph neural network (GNN) and Multilayer perceptron (MLP) are 
 popular amplification of function $f(\cdot)$. Considering a 
 specification of noise ($E$) distribution sampling 
 $\{H_{s}\}^{S}_{s=1}$ in a dataset with $S$ dialogue samples, 
 Equation~\ref{eqt5} can be written by a maximization of log-evidence: 

 \begin{equation}
  \frac{1}{S} \sum_{s=1}^{S} \log p(H_{s})=\frac{1}{S} \sum_{s=1}^{S} \log\int p(H_{s}|E)p(E)dE
  \label{eqt6}
  \end{equation}

Continuing the theory of variational Bayes, we regard $E$ as 
the latent variable in variational autoencoder (VAE)~\cite{kingma2022autoencoding} 
and use variational posterior $q(E|H)$ to approximate 
the intractable posterior $p(E|H)$ (See details in Figure~\ref{figPF}), 
thus the evidence lower bound (ELBO) reads: 

\begin{equation}
  \mathcal{L}^{s}_{ELBO}=-KL(q(E|H_{s})||p(E))+E_{q(E|H_{s})}[\log p(H_{s}|E)]
  \label{eqt7}
  \end{equation}

From the VAE view, for each dialogue embedding $H_{s}$, 
the inference model encodes it to output latent variable $E$ 
with $q(E|H_{s})$, and then the decoder 
reconstructs $H_{s}$ from $E$ via $p(H_{s}|E)$. 
From the causal graph view, defining that $\mathcal{H}$ 
represents the domain of dialogues: 
$\{H_s\}^{S}_{s=1} \in \mathcal{H}$, $\mathcal{E}$ represents the 
domain of exogenous variable $E$, $\widehat{\mathcal{H}}$ represents 
the domain of causal representation of $\mathcal{H}$. 
Therefore, a variational model consists of two functions: 
an exogenous variable inference 
(also called implicit causes inference): 
$\mathcal{H} \rightarrow \mathcal{E}$, and observation relation 
reconstruction: $\mathcal{E} \rightarrow \widehat{\mathcal{H}}$.

\subsection{Probabilistic Implementation}

\subsubsection{Encoder}

The encoder $q_{\varphi}(z|\mathcal{X})$ applies a graph attention 
module $f_{att,\varphi}$~\cite{velivckovic2017graph} to the input. 

\begin{equation}
  q_{\varphi}(z|\mathcal{X})=softmax(f_{att,\varphi}(X))
  \label{eqt8}
  \end{equation}

Specifically, It produces an adjacent matrix $A^{\ell}$,  
$\ell={1, 2, \dots, L-1}$ represents the layer of GNN. 
Thus, for each utterance embedding $h^{\ell}$ at the $\ell$-th layer, 
the $A^{\ell}_{i,t}$ computed by attention mechanism is a weighted combination 
of $h^{\ell}_{t}$ for each directly related utterance 
$h^{ell}_{i} (i\in rel_{t})$: 

\begin{equation}
  A^{\ell}_{i,t}=\frac{LeakyReLU(e_{i,t}^{\ell})}{\sum_{j\in rel_{t}}LeakyReLU(e_{j,t}^{\ell})} 
\label{eqn9}
\end{equation}

\begin{equation}
  e_{i,t}^{\ell}=\overrightarrow{h}_{i}W^{\ell}_{i(row)}+(\overrightarrow{h}_{t}W^{\ell}_{t(col)})^{T}
  \label{eqn10}
\end{equation}
where $W^{\ell}_{row}\in \mathbb{R}^{N\times1}$ and $W^{\ell}_{col}\in \mathbb{R}^{N\times1}$ 
are the learnable parameters in the graph attention. Moreover, the 
GNN aggregates the information from the neighbor utterances as follows: 

\begin{equation}
  H^{\ell+1}=eLU((I-(A^{\ell})^{T})H^{\ell}W^{\ell})
  \label{eqn11}
\end{equation}
where $W^{\ell}$ stands for parameters in the corresponding layer. From the 
final layer of the evaluation process, 
by extracting $A^{L-1}$ computed in Equation~\ref{eqn9}, 
the marginal or conditional ``distribution'' of $H$ is obtained.
Besides, by extracting $H^{L}$ in Equation~\ref{eqn11}, we can 
have the substitute for the implicit causes $E=MLP(H^{L})$. 

We introduced a Gumbel distributed noise  $\varepsilon$~\cite{maddison2016concrete} 
to be capable of backpropagating via the discrete distribution samples. 
\begin{equation}
  z \sim softmax(E+\varepsilon)
  \label{eqt28}
  \end{equation}

The output $z$ implies the possible distribution of implicit causes 
over $\mathcal{X}$. In this cohesive way, our 
method addresses the challenge of causal discovery in the conversation.

\subsubsection{Decoder}

The decoder accumulated the incoming messages to each node via 
latent variable $z$ and employed a new graph neural network 
$f_{gnn,\theta}$: 

\begin{align}
  p_{\theta}((x|z^{-1}E))=f_{gnn,\theta}((I-A)^{-1},z)
\label{eqt30}
\end{align}

Specifically, With a fixed adjacency matrix, 
the GNN aggregates the information of implicit causes 
from neighbor nodes as follows: 

\begin{equation}
  \widehat{E}^{\ell+1}=eLU((I-(A^{L})^{T})^{-1}E^{\ell}M^{\ell})
  \label{eqn9}
\end{equation}

where $M^{\ell}$ is parameters in the corresponding layer, 
$E^{\ell}$ represents the embedding of implicit causes in $\ell$-th 
layer, and $E^{1}=z$. As the same 
architecture as the encoder, $\widehat{H}=MLP(E^{L})$.

The output of the decoder $\widehat{H} \in \mathbb{R}^{N \times D}$ equals 
the dimension of $\mathcal{H}$ and it is the causal representation 
of $U$ reconstructed by implicit causes $E$. 

\subsubsection{Evidence Lower Bound}

Our variational lower bound consists of a reconstruction error 
$l_{rc}$ measuring the distance of $\widehat{H}$ and $H$, and 
a KL divergence measuring the prior probability $P(E)$ and variational 
posterior $q_{\varphi}(z|\mathcal{X})$

\begin{align}
  \mathcal{L}=&l_{rc}(H,\widehat{H})-KL[q_{\varphi}(z|\mathcal{H})||\mathbb{P}(z)]\\
    =&E_{q_{\varphi}(z|\mathcal{H})}[l_{rc}(H,f_{\theta}((I-A)^{-1}z)]-KL[q_{\varphi}(z|\mathcal{H})||\mathbb{P}(z)]
  \label{eqt34}
  \end{align} 
where $P(E)$ is the standard matrix normal: 
$p(E)=\mathcal{M}\mathcal{N}_{N \times D}(0,I,I)$, 
$l_{rc}$ was adopted as mean squared error (MSE): 

\begin{equation}
  l_{rc}(H,f_{\theta}((I-A)^{-1}z)=\frac{1}{D} \sum_{i = 1}^{D} (H_{i}-\widehat{H}_{i})^{2}
\end{equation}

In addition, for different downstream tasks, 
there are different auxiliary loss functions to constrain $\widehat{H}$.
Generally, these tasks require a classifier to 
transform $\widehat{H}$ into the corresponding predicted results, 
and then measure the error with the ground truth. 

\section{Dataset}

\subsection{Overview}

\begin{table}
  \footnotesize
  \centering
  \begin{threeparttable}   
  \resizebox{\linewidth}{!}{
    \begin{tabular}{|c|c|c|c|c|c|c|}
      \hline
    \multirow{2}{*}{Dataset}&\multirow{2}{*}{Data Type}&\multicolumn{3}{c}{Statistics}\vline&\multicolumn{2}{c}{Causes Label}\vline\\
    \cline{3-7}
    &&Train&Eval&Test&Explicit&Implicit\\
    \hline
    RECCON&dialogue&833&47&225&$\bigtriangleup$ \tnote{1} &-\\
    Synthetic&single value&833&47&225&$\bigtriangleup$ &$\surd$ \tnote{2} \\
    Simulation&dialogue&1381&100&200&$\surd$ &$\bigtriangleup$ \\
    \hline
  \end{tabular}}
  \begin{tablenotes}    
    \footnotesize               
    \item[1] $\bigtriangleup$ represents this dataset has partial indicators of the label. 
    \item[2] $\surd$ represents this dataset has full indicators of the label.
  \end{tablenotes}            
\end{threeparttable}       
  \caption{Statistics of Synthetic and Real-world datasets}
  \label{tabdataset}
\end{table}

One of the reasons for the insufficient substantial research on 
conversation reasoning, in addition to the lack of an algebraic 
cognitive model, is the absence of an appropriate dataset. 
Among the existing data resources, only the RECCON~\cite{poria2021recognizing}
dataset contains partial utterance-level causal indicators. 
However, for a complete dialogue cognitive model, 
research on implicit causes is actually necessary. 
Implicit causes, stemming from Section~\ref{sec3.1}, 
represent unobservable components of a speaker's unique experiences, 
personal desires, etc., which even in real-world datasets, 
are difficult for creators to collect 
from the observer's perspective. 

To enable quantitative evaluation 
of affective reasoning, particularly with respect to implicit causes, 
we propose a synthetic dataset and a simulation dataset. 
The synthetic dataset adopts a DAG structure 
from the causal discovery field, where each variable is 
generated according to the SCM generating process (i.e., Equation~\ref{eqn1}), 
wherein each variable contains explicit and implicit cause data. 
The simulation dataset, on the other hand, is a corpus of 
short dialogues generated by the large language model GPT-4. 
In the generation process, we control the knowledge 
of the preceding sentence to ensure that the cause of 
the target utterance aligns with our design. 
Table ~\ref{tabdataset} illustrates the similarities and differences 
between the synthetic, simulation, and real-world datasets. 
Overall, the synthetic and simulation datasets 
effectively alleviate the lack of crucial information 
in datasets for conversation reasoning. 

\subsection{Real-World and Synthetic Dataset}

\textbf{RECCON}: 
The first dataset for emotion-cause recognition of conversation 
including RECCON-DD and RECCON-IE (emulating an out-of-distribution 
generalization test). RECCON-DD includes 5380 labeled ECPs and 5 cause 
spans (\textit{no-context}, \textit{inter-personal}, \textit{self-contagion}, 
\textit{hybrid}, and \textit{latent}). For utterances with 
emotional tendencies, explicit cause indicators 
(i.e., one or more preceding utterances) are annotated, 
while most non-emotional utterances are not annotated. 

\textbf{Synthetic dataset}: 
We create a synthetic dataset by following the benchmark of the 
causal discovery field~\cite{agrawal2021decamfounder,squires2022causal}. 
To minimize sample bias, we did not randomly draw causal graphs as 
samples. Inversely, the number of samples in the synthetic dataset 
and the number of utterances and labels per sample are restricted to be consistent with RECCON.
We use Causal Additive Models (CAMs), Specifically SCM structure for our datasets. 
First, we assume that each $i.i.d.$ implicit causes
$E \sim \|^{50}  \mathcal{N} (1,1)$ if it is an emotion utterance in the original dataset, 
and $E \sim \|^{50} \mathcal{N} (-1,1)$ if it is not. Then, we update each utterance via speaker turns $S$: 
if there is a emotion-cause pair  $(U_{i},U_{j})\in L $, 
then $U_{i}=\alpha_{j,i} U_{j}+E_{i}$ ($\alpha_{j,i}\sim Unifrom ([0.7,1])$), 
and for those pairs without emotion-cause label, $\alpha_{j,i}\sim Unifrom ([0,0.3])$. 
Finally, we randomly select a noise $\xi \sim Unifrom ([-0.25,0.25])$ for each 
utterance $U_{i}=U_{i}+\xi_{i}$. 

\begin{figure}
  \centering
  \subfloat[Chain\_I]{
    \includegraphics[width=0.24\linewidth]{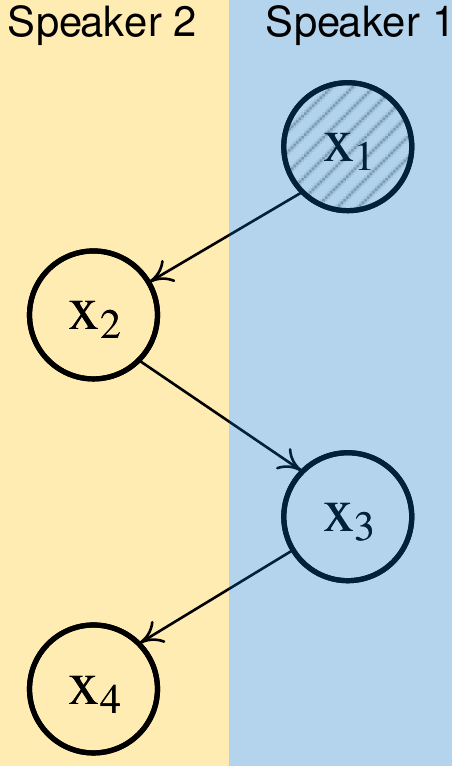}}
  \subfloat[Chain\_II]{
    \includegraphics[width=0.24\linewidth]{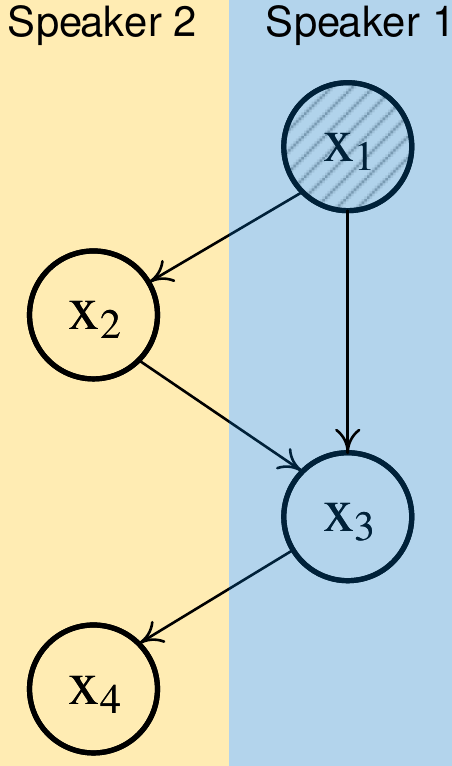}}
 \subfloat[Chain\_III]{
    \includegraphics[width=0.24\linewidth]{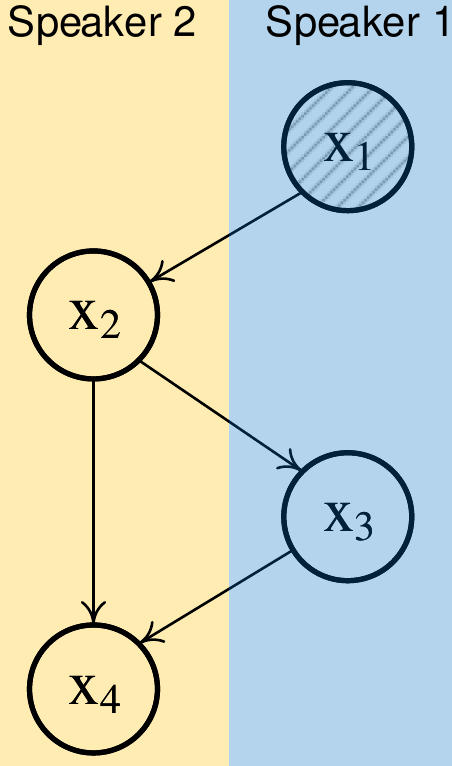}}
  \subfloat[Chain\_IV]{
    \includegraphics[width=0.24\linewidth]{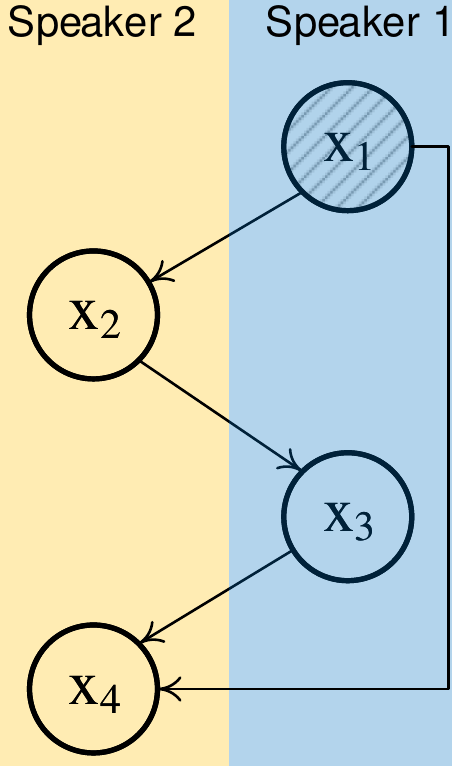}}
  \caption{Four causal structures in the simulation dataset.}
  \label{figchain}
\end{figure}

\subsection{Simulation Dataset}

we consider GPT-4~\cite{openai2023gpt4} to generate dialogues according to designed 
causal structures. Specifically, we for simplicity assumed all 
dialogues are composed of 4 utterances rotating from 2 speakers, 
and designed 4 causal structures to manifest different causal 
relationships. Along with the designed generating rules according 
to 4 causal structures, we adopted gpt-4-0314
\footnote{https://platform.openai.com/docs/models/gpt-4} 
and Chat Completion api
\footnote{https://platform.openai.com/docs/api-reference/chat} to 
commence dialogue generation. We began with the introduction of 4 
causal structures to elaborate our building process (See Figure~\ref{figchain}): 

\textbf{Chain\_I} is a fundamental structure in dialogue, in which 
causal relationship is consistent with speaking turns. Specifically, 
Chain\_I has and only has $x_{1}\rightarrow x_{2}$, 
$x_{2}\rightarrow x_{3}$, 
and $x_{3}\rightarrow x_{4}$, where $x$ represents the utterance. 

\textbf{Chain\_II} is a variation of Chain\_I. Besides 
$x_{1}\rightarrow x_{2}$, $x_{2}\rightarrow x_{3}$, 
and $x_{3}\rightarrow x_{4}$, 
Chain\_II additionally has $x_{1} \rightarrow x_{3}$. 

\textbf{Chain\_III} is a variation of Chain\_I. Besides 
$x_{1}\rightarrow x_{2}$, $x_{2}\rightarrow x_{3}$, 
and $x_{3}\rightarrow x_{4}$, 
Chain\_III additionally has $x_{2} \rightarrow x_{4}$. 

\textbf{Chain\_IV} is a variation of Chain\_I. Besides 
$x_{1}\rightarrow x_{2}$, $x_{2}\rightarrow x_{3}$, 
and $x_{3}\rightarrow x_{4}$, 
Chain\_IV additionally has $x_{1} \rightarrow x_{4}$. 

In the generation process, 
we set the \textit{system} to be the same for the same person, 
while the content of the \textit{user} is determined 
based on the specific skeleton
\footnote{In early causality discovery algorithms, 
the ``causal skeleton'' was used to represent a partial directed or 
undirected graph that contains prior relationships. 
In our dataset, the ``causal skeleton'' is employed to represent 
the prior-set causal structure, thus equal to the ``causal structure''}. 
Below is an example dialogue 
of the Chain$\_$IV from our dataset: 

\textit{1. Have you seen a dog around here recently?}

\textit{2. I'm sorry, but I haven't seen any dogs around here lately.}

\textit{3. That's okay, thank you for your time.}

\textit{4. Excuse me, do you mind if I ask why you are looking for a dog?}

The first and third utterances are attributed to the same speaker, 
while the second and fourth utterances share the same speaker identity as well. 
Their \textit{system} is set to \textit{
  \{``system$\_$speaker1'': ``You are a policeman, received helpers said her dog lost, her dog is a white Samoye.''
  ``system$\_$speaker2'': ``You are a taxi driver. You saw someone took a dog in your taxi car last night.''\}
}
Furthermore, when generating the third utterance, 
we only set 
\textit{\{``role'': ``system", ``content": system$\_$speaker1\},
\{``role": ``user", ``content": utterance$\_$2\}} 
according to the Chain$\_$IV protocol. 
However, when generating the fourth utterance, we set 
\textit{\{``role'': ``system", ``content": system$\_$speaker2\},
\{``role": ``user", ``content": utterance$\_$1\},
\{``role": ``user", ``content": utterance$\_$3\}}. 

However, this dataset also confronts several limitations 
and biases. Firstly, our construction only covers ``chain'' 
structures, neglecting another equally important 
``fork'' structures (e.g., $x_{1}\rightarrow x_{2}, 
x_{1}\rightarrow x_{3}, x_{1}\rightarrow x_{4}$), 
as sceneries featuring the ``fork'' are typically 
more uncommon in dialogues. Additionally, 
the absence of manual verification leads to the prevalence 
of pervassive responses (for instance, \textit{``thanks''}) 
in the samples. This could render the involved causal relationships 
more obscure and harder to discriminate. While manual verification 
could enhance the distinctiveness between structures, 
it might introduce biases such as annotation consistency. 
Therefore, a future version of this dataset should encompass 
chain, fork, and hybrid causal structures, 
together with clear samples from high-quality manual verification. 

Furthermore, there are concerns regarding GPT-4's inherent biases - 
are GPT-4's responses similar to human responses? 
The impact of these inherent biases is particularly pronounced 
in dialogue generation tasks. Thus, we suggest extra considerations 
are needed when this dataset is used for generation tasks. 
However, in the tasks presented in this paper, 
the effect of inherent bias is little to nonexistent. 
The bias we are concerned about is whether the utterance is 
smooth and fluid, and in that regard, GPT-4 performs very well. 

\section{Experiments} 
\subsection{Tasks and Metrics}
We incorporated two distinctive tasks in our study: 
Explicit Cause Extraction (ECE) and 
Implicit Cause Extraction (ICE) tasks. The ECE task was 
implemented to demonstrate the effectiveness of our model, 
whereas the ICE task aimed at contributing toward explainability. 

More specifically, the ECE task is derivative of the 
existing ECPE task~\cite{xia-ding-2019-emotion}, 
that seeks to determine whether there exists a causal 
relationship $U_{j}\rightarrow U_{i}$ for any pair of utterances 
$(U_{i}, U_{j})$. Due to the limitations in the labels present 
in the real-world and synthetic datasets, we were only able to 
evaluation the pairs where $U_{i}$ is emotional utterance. 
However, for the simulated dataset, we can examine 
all utterances pairs.  

For the ICE task, we tested the sentiment 
consistency between the implicit causes and the corresponding 
utterances, i.e., whether $E_{i}$ and $H_{i}$ have the same 
emotion prediction, which can also be 
theoretically proven by Equation~\ref{eqt5}. Assume that 
$W=(I-A)^{-1}$ and $A_{i,i}=0$, so in $W$, 
the value of the elements on the diagonal is constant at $1$ and 
is a constant maximum of each column. 
Naturally, $f(E)$ is an approximate estimate of $H$. Hence, we 
only conduct the ICE task in real-world and synthetic datasets 
(simulation dataset does not labeled with emotion categories). 

We used $F1$ score as the evaluation metric for both tasks. 
The $F1$ score of the explicit task indicates the model's ability 
to identify explicit causes, while the $F1$ score of the 
implicit task indicates the model's interpretability of 
implicit causes (We consider that the implicit causes are 
interpretable when the $F1$ score of the implicit task 
is greater than $80$, indicating sentiment consistency 
between the implicit causes and the corresponding utterances.). 

\subsection{Baselines}

As dialogue reasoning is still in its infancy in academic research, 
we consider the following methods as baselines for dialogue reasoning. 

\textbf{Transformer-based models}:
\textbf{RoBERTa}~\cite{liu2019roberta}, is widely 
regarded as state-of-the-art benchmarks 
for many downstream text tasks. 

\textbf{GNN-based models}:
We adopted several baseline models that have demonstrated 
excellent performance in the field of affective reasoning, 
including 
\textbf{EGAT}~\cite{chen2022learning}, 
\textbf{DECN}~\cite{lian2021decn}, 
\textbf{DAG-ERC}~\cite{shen-etal-2021-directed},
and \textbf{CAE}~\cite{chen2023affective}. 
Their detailed descriptions can be found in~\cite{chen2023affective}.

\textbf{Large Language Model}: In addition, 
we included \textbf{GPT-3.5} and \textbf{GPT-4}\cite{openai2023gpt4} 
as our baseline models, which have achieved state-of-the-art 
performance on most downstream text tasks 
surpassing supervised learning. 

\subsection{Implementation Details}
In the word embedding, we adopt the affect-based pre-trained features
\footnote{\url{https://drive.google.com/file/d/1R5K_2PlZ3p3RFQ1Ycgmo3TgxvYBzptQG/view?usp=sharing}} 
proposed by~\cite{shen-etal-2021-directed} for experiments in the RECCON dataset and 
Roberta-base
\footnote{\url{https://huggingface.co/roberta-base}}
 for the simulation dataset. Our code is open in GitHub\footnote{\url{https://github.com/Zodiark-ch/masters-of-our-EMNLP2023-papers}}.

 In the hyper-parameters, we follow the setting of~\cite{chen2023affective}. 
Specifically, the learning rate is set to 3e-5, 
batch size is set to 32, and epoch is set to 60. Furthermore, we set $L$ 
to 1, and implicit cause size is set to 192, hidden size of GNN is set to 
300, and dropout rate is 0.3. 

Regarding the hyper-parameters of the baselines, 
we followed the settings of~\cite{chen2023affective}, 
such as setting the learning rate at 3e-5, batch size at 32, 
epoch at 60, $L$ at 1, 
GNN hidden size at 300. The complete parameter settings 
can be found in~\cite{chen2023affective}. 

Meanwhile, due to the insufficient size of testset in all three datasets, we 
evaluated our method ten times with different data splits by following~\cite{chen2023affective} 
and then performed paired sample $t$-test on the experimental results.

\subsection{Results in Real-World and Synthetic Datasets}
\label{secrrs}

\begin{table}
  \footnotesize
  \centering
  \begin{threeparttable}   
  \resizebox{\linewidth}{!}{
  \begin{tabular}{|c|c|c|c|c|}
    \hline
    \multirow{2}{*}{model}&\multicolumn{2}{c}{Real-world}\vline&\multicolumn{2}{c}{Synthetic}\vline\\
    \cline{2-5}
    &explicit\tnote{1}&implicit\tnote{2}&explicit&implicit\\
    \hline
    GPT-3.5&47.2$\pm$2.1&-&-&-\\
    GPT-4&51.1$\pm$2.7&-&-&-\\
    \hline
    RoBERTa&62.71$\pm$1.8&-&74.51$\pm$0.9&-\\
    RoBERTa$^{+}$\tnote{3}&63.84$\pm$1.5&-&74.83$\pm$1.2&-\\
    \hline
    EGAT&68.05$\pm$1.5&82.19$\pm$2.8&72.59$\pm$4.8&82.49$\pm$2.7\\
    DECN&68.32$\pm$1.5&-&78.42$\pm$1.9&-\\
    DAG-ERC&70.36$\pm$1.5&-&79.92$\pm$1.4&-\\
    CAE$_{1}$\tnote{4}&69.16$\pm$1.2&92.83$\pm$1.9&80.94$\pm$1.9&91.93$\pm$1.5\\
    CAE$_{2}$&70.12$\pm$2.1&93.89$\pm$2.7&81.29$\pm$1.5&93.54$\pm$1.4\\
    CAE$_{3}$&73.17$\pm$1.1&94.16$\pm$2.4&82.11$\pm$2.9&94.29$\pm$1.6\\
    \hline
    Ours&\textbf{77.59$\pm$1.5}&\textbf{96.31$\pm$2.4}&\textbf{85.46$\pm$1.7}&\textbf{97.13$\pm$2.1}\\
    \hline
    
  \end{tabular}}
  \begin{tablenotes}    
    \footnotesize               
    \item[1] represents the ECE task. 
    \item[2] represents the ICE task.
    \item[3] represents the large version of RoBERTa (1024 dimensions). 
    \item[4]We adopted 3 CAE frameworks, CAE$_{1}$, CAE$_{2}$, CAE$_{3}$ 
    correspond to Skeleton III, Skeleton IV, and Skeleton VI, respectively, 
    in Paper~\cite{chen2023affective}.
  \end{tablenotes}     
\end{threeparttable} 
  \caption{Overall performance in Real-World and Synthetic datasets.}
  \label{taberands}
\end{table}

We conducted ECE and ICE tasks on 
both real and synthetic datasets, as shown in Table~\ref{taberands}. 
Furthermore, due to the lack of visibility of 
intermediate representations and the limited ability 
to handle single-value data in the GPT series networks, 
we could only conduct explicit evaluations on real datasets 
for them. Specifically, we augmented each sample input to GPT 
with a dialogue sample containing five utterances 
and 2 emotion-causes pairs, using prompts. 

\subsubsection{ECE task}

Our method outperformed other methods significantly in 
both real-world and synthetic datasets, 
indicating that effective cognitive model-driven approaches 
can better capture reasoning and learning between utterances. 
At the same time, this philosophy can also be demonstrated 
by the 3 results of CAE. Among the 3 results of CAE, 
CAE$_{3}$ is the causal discovery method 
closest to our cognitive model structure, 
and therefore performs second best to our method. 
In addition, both our method and CAE are designed 
with causal discovery, including the causal strength matrix 
for the effect of utterances and the autoregression 
of exogenous variables, which is significantly better than 
other non-causal discovery methods, such as DAG-ERC's 
directed graph learning, DECN's correlation graph learning, 
EGAT's enhanced graph learning, and other apparent 
graph relationship learning methods. 

\subsubsection{ICE task}

we tested methods that can represent implicit causes, 
including EGAT, CAE, and our method. 
All methods involved in this test were able to satisfy 
the sentiment consistency condition of implicit causes 
($F1$ score greater than 80), which experimentally confirms 
our theoretical work on sentiment consistency. 
However, sentiment consistency is only an approximate estimate, 
so we have emphasized that the purpose of implicit evaluation tasks 
is to demonstrate the interpretability of implicit causes. 
In other words, as long as the sentiment consistency condition is met, 
the higher $F1$ score is difficult to demonstrate better performance, 
though our method performs the best. 

\begin{table}
  \footnotesize
  \centering
  \resizebox{0.9\linewidth}{!}{
  \begin{tabular}{|c|c|c|c|c|}
    \hline
    \multirow{2}{*}{Strategies}&\multicolumn{2}{c}{Real-world}\vline&\multicolumn{2}{c}{Synthetic}\vline\\
    \cline{2-5}
    &explicit&implicit&explicit&implicit\\
    \hline
    Ours&77.59&96.31&85.46&97.13\\
    \hline
    Linear$_{d}$&$\downarrow$0.62&$\downarrow$0.74&$\downarrow$0.45&$\downarrow$0.66\\
    Linear$_{e}$&$\downarrow$0.77&$\downarrow$0.95&$\downarrow$0.72&$\downarrow$1.07\\
    GAT$_{d}$&$\downarrow$1.79&$\downarrow$0.79&$\downarrow$1.69&$\downarrow$0.83\\
    GNN$_{e}$&$\downarrow$1.85&$\downarrow$1.39&$\downarrow$1.92&$\downarrow$1.22\\
    Sample&$\downarrow$2.11&$\downarrow$3.59&$\downarrow$1.97&$\downarrow$3.63\\
    $-E$&$\downarrow$4.16&-&$\downarrow$5.28&-\\
    \hline
  \end{tabular}}
  \caption{Ablation results}
  \label{tabablation}
\end{table}

\subsubsection{Ablation Study}

To thoroughly investigate which component contributes the most 
to our model, we conducted ablation experiments as shown in Table~\ref{tabablation}. 
We performed the following four operations on the network module: 
Linear$_{d}$: replacing the GNN in the decoder with Linear layer; 
Linear$_{e}$: replacing the GAT in the encoder with Linear layer; 
GAT$_{d}$: replacing the GNN in the decoder with GAT; 
GNN$_{e}$: replacing the GAT in the encoder with GNN. 
Meanwhile, we performed two operations on the network structure: 
Sample: removing the sampling on latent variable and Loss$_{KL}$; 
-E: removing the decoder (i.e., 
adopting the outputs of encoder as final representation). 

The results from Linear$_{d}$ and Linear$_{e}$ indicate that 
our model is not sensitive to the type of network module, 
as demonstrated on both explicit and implicit tasks. 
However, under the GAT$_{d}$ and GNN$_{e}$ operations, 
there was a significant decrease in performance. GAT$_{d}$ 
means that the decoder does not inherit the $A$ matrix 
calculated in the encoder, but instead calculates a new matrix 
to replace $(I-A)^{-1}$. GNN$_{e}$ represents that encoder does not 
calculate the dynamic $A$ matrix, but instead adopts the fixed 
mask matrix introduced in~\cite{chen2023affective}. 
Both of these two operations affect the design of 
the causal strength matrix $A$ and disrupt our reasoning process, 
so it is natural that they result in a significant decrease 
in performance. Finally, Sample and -E essentially disrupt 
the variational inference, making the network model 
unable to correspond to the SCM and resulting 
in the most severe decrease in performance. 

\begin{figure*}
  \subfloat[E at epoch1]{
    \includegraphics[width=0.19\linewidth]{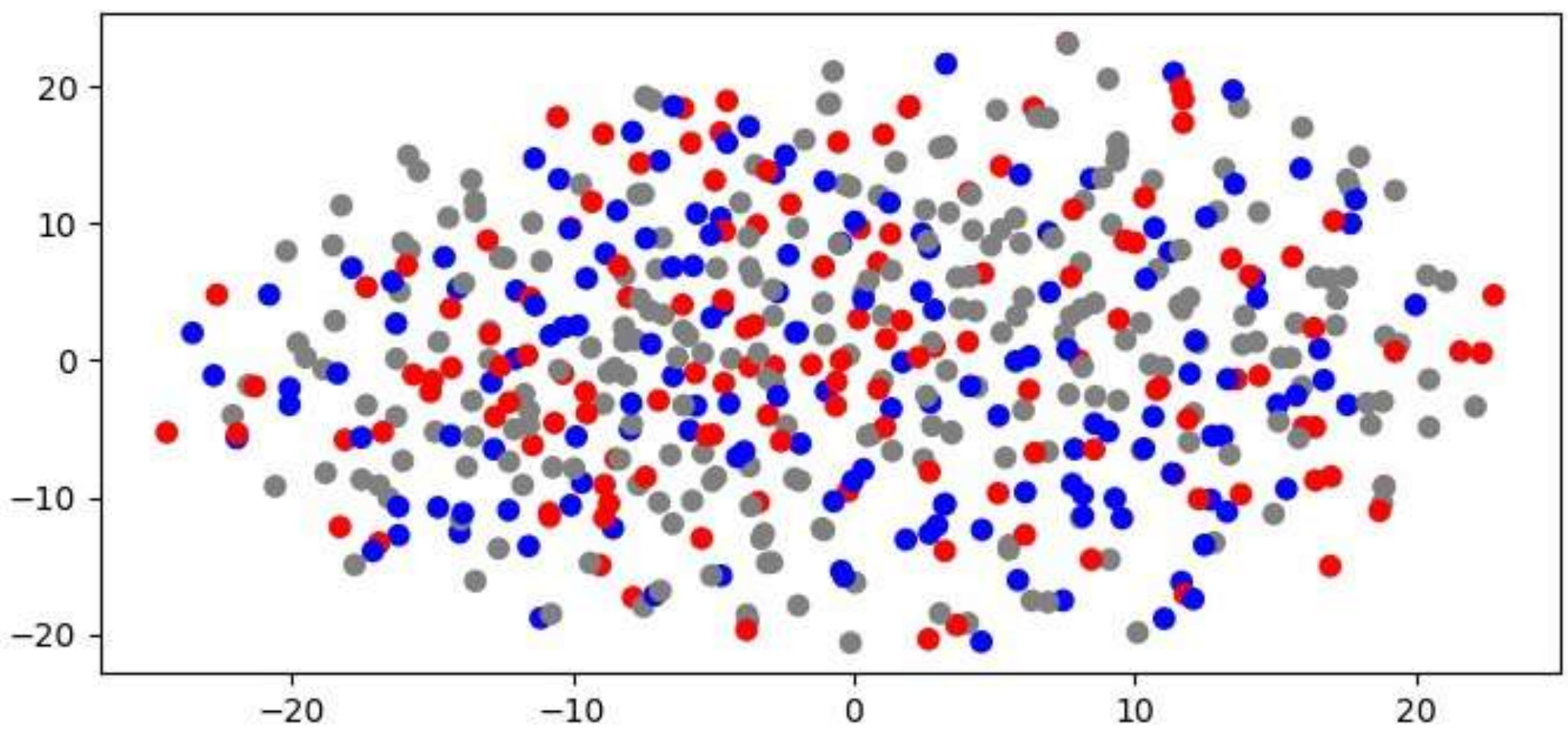}}
  \subfloat[E at epoch10]{
    \includegraphics[width=0.19\linewidth]{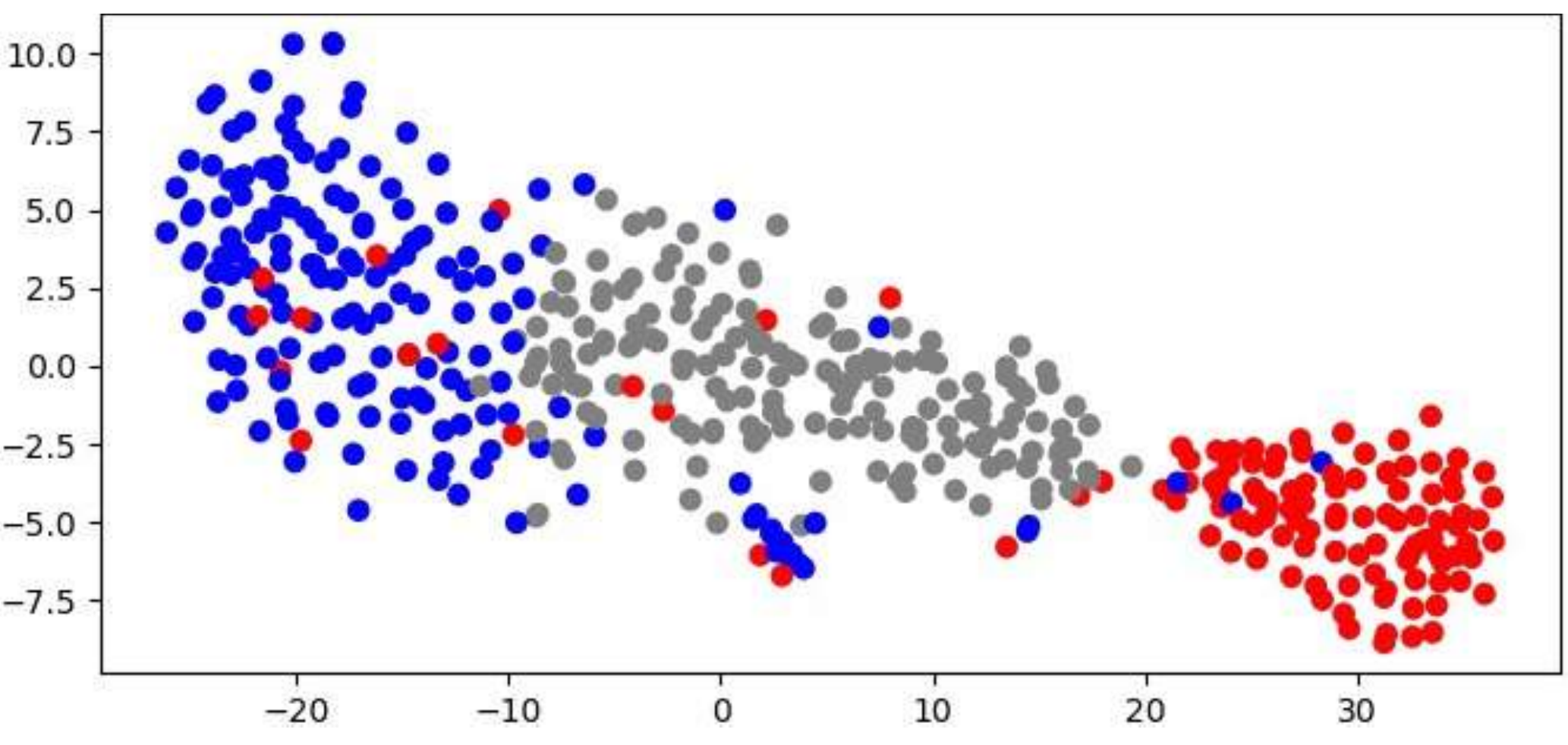}}
  \subfloat[E at epoch40]{
    \includegraphics[width=0.19\linewidth]{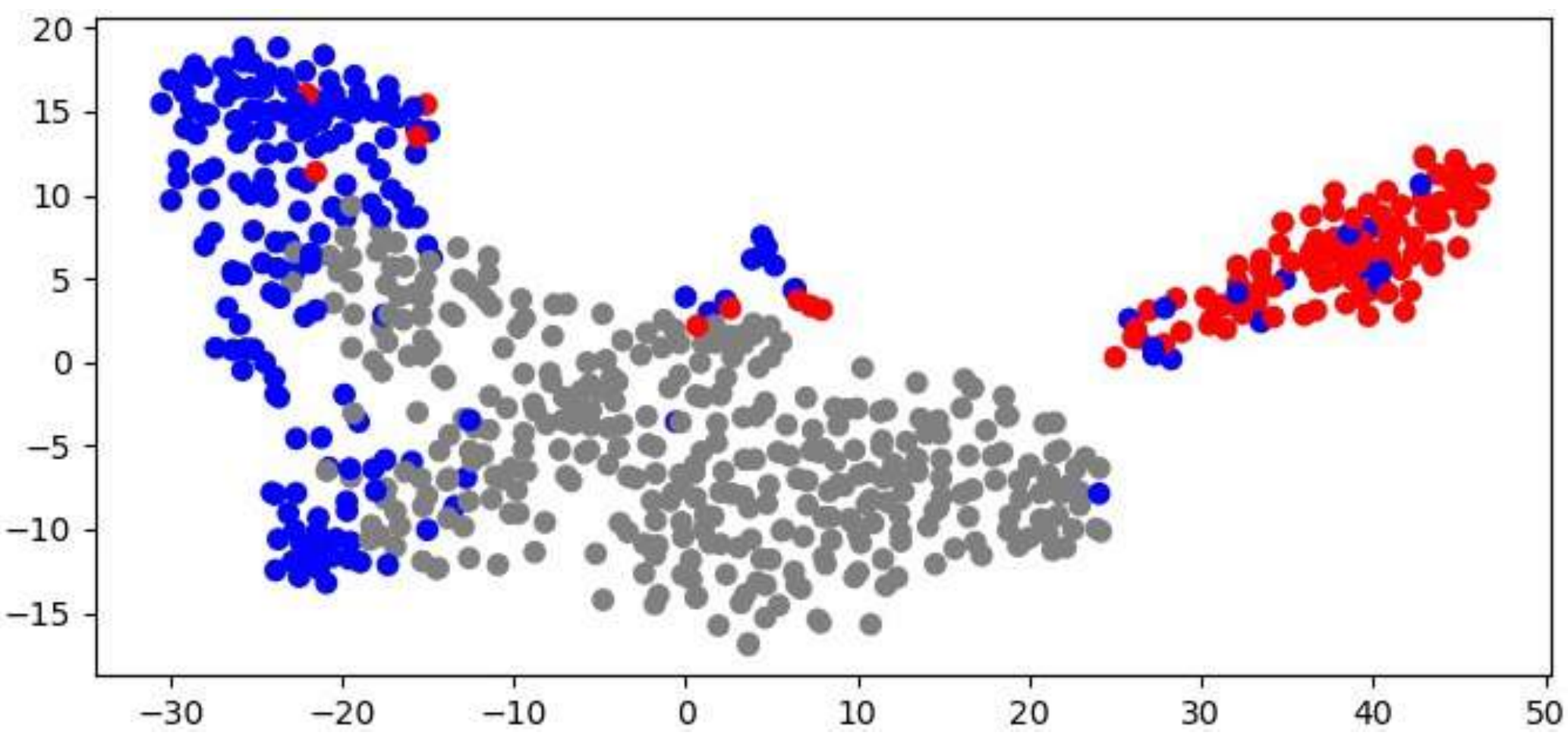}}
  \subfloat[E at epoch60]{
    \includegraphics[width=0.19\linewidth]{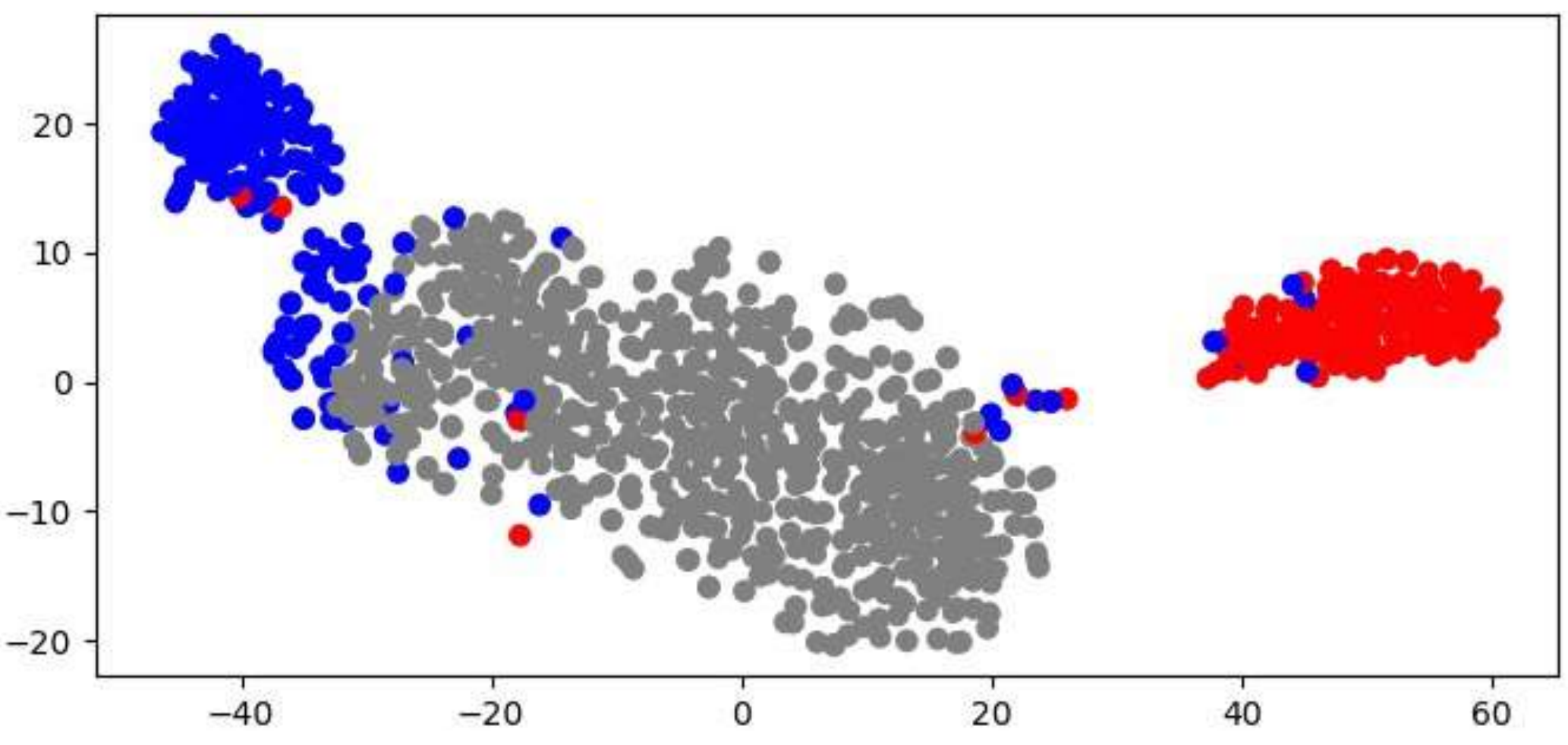}}
  \subfloat[true sample]{
    \includegraphics[width=0.19\linewidth]{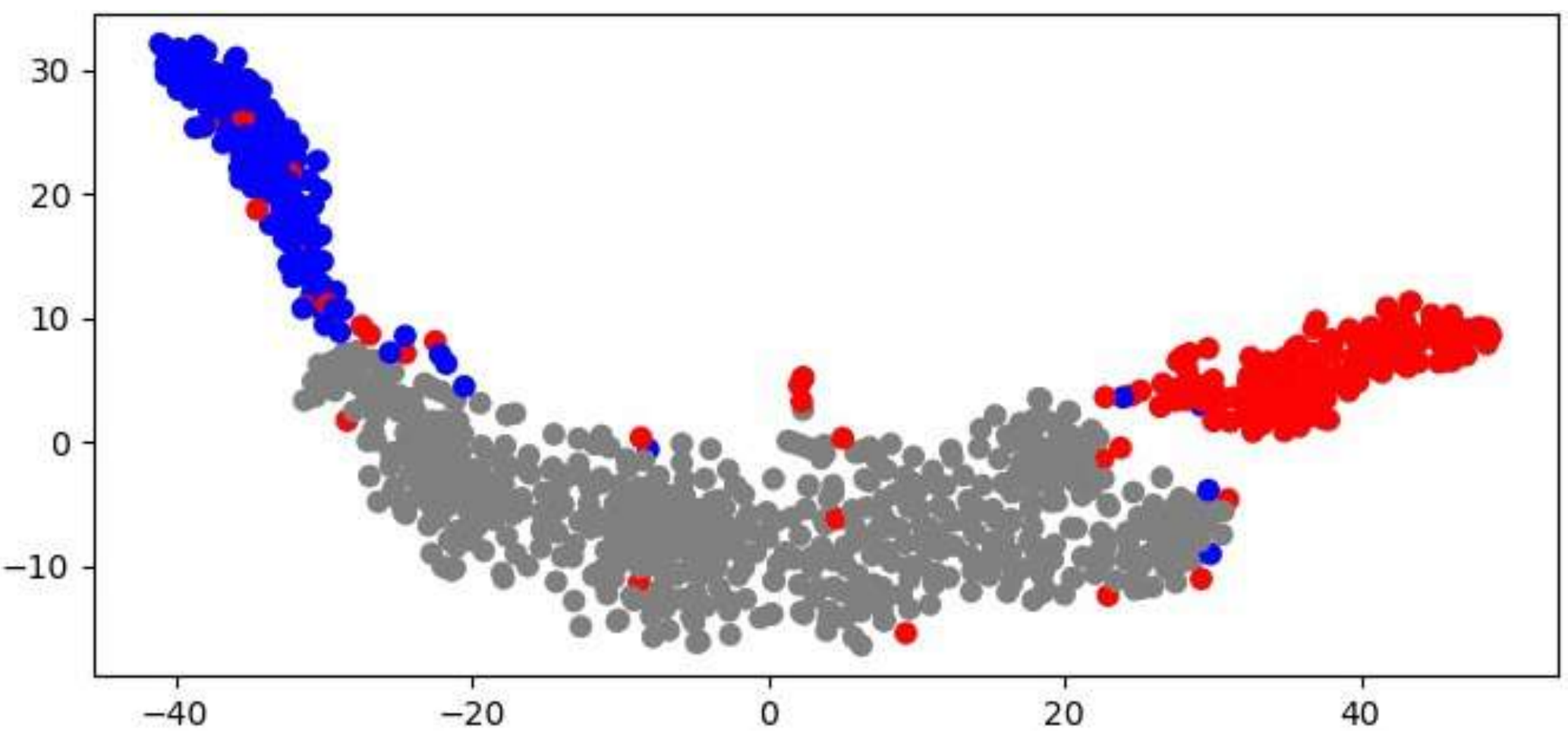}}
  \caption{Visualization of $E$ (a-d subfigures) and 
  implicit causes (e subfigure) with colors in the 
  synthetic datasets. The gray cluster means padding utterances 
  in each dialogue, the blue cluster corresponds to the non-emotion utterances, 
  and the red cluster corresponds to emotion utterances. 
   }
  \label{figlatent}
\end{figure*}
\subsubsection{Visualization}
Furthermore, Figure~\ref{figlatent} shows the projection 
of $E$(a-d) and ground truth of implicit causes (e), respectively, using 
t-SNE~\cite{knyazev2019understanding} on the synthetic dataset. 
We observe that $E$ and implicit causes are similarly clustered 
into three parts when epoch is 60 which indicates that E is consistent 
with the implicit causes in the samples through the distribution. 
Finally, the whole process of learning (from epoch1 to epoch 60) 
indicates that $E$ successfully learns the implicit causes.

\subsection{Results in Simulation Dataset}

\begin{table}
  \footnotesize
  \centering
 
  \resizebox{\linewidth}{!}{
  \begin{tabular}{|c|c|c|c|c|}
    \hline
    \multirow{2}{*}{model}&\multicolumn{4}{c}{Skeleton}\vline\\
    \cline{2-5}
    &I&II&III&IV\\
    \hline
    GPT-3.5&52.14$\pm$4.3&57.22$\pm$4.7&53.94$\pm$3.5&51.51$\pm$6.2\\
    GPT-4&57.3$\pm$5.8&61.84$\pm$4.4&55.45$\pm$3.6&51.65$\pm$5.6\\
    \hline
    RoBERTa&61.46$\pm$4.5&63.34$\pm$3.6&63.46$\pm$3.5&60.38$\pm$4.7\\
    RoBERTa$^{+}$&61.78$\pm$4.1&62.87$\pm$4.4&62.78$\pm$3.8&60.18$\pm$4.8\\
    \hline
    EGAT&63.48$\pm$4.5&64.97$\pm$3.2&62.78$\pm$4.7&62.78$\pm$5.7\\
    DECN&63.86$\pm$4.3&63.81$\pm$3.4&61.18$\pm$4.9&60.91$\pm$3.7\\
    DAG-ERC&65.85$\pm$3.8&61.37$\pm$3.7&64.18$\pm$4.6&59.74$\pm$4.4\\
    CAE$_{1}$&64.18$\pm$5.1&65.48$\pm$3.9&62.18$\pm$4.3&61.18$\pm$4.3\\
    CAE$_{2}$&65.97$\pm$3.7&65.48$\pm$4.1&63.15$\pm$7.1&61.29$\pm$4.4\\
    CAE$_{3}$&66.48$\pm$4.5&64.19$\pm$3.4&62.94$\pm$3.9&62.48$\pm$3.9\\
    \hline
    Ours&\textbf{69.99$\pm$3.5}&\textbf{71.49$\pm$3.4}&\textbf{71.89$\pm$4.2}&\textbf{68.11$\pm$4.4}\\
    \hline
    
  \end{tabular}}

  \caption{Overall performance in Simulation datasets.}
  \label{tabes}
\end{table}

On Simulation dataset, we also conducted ECE task 
as we had complete pair labels on Simulated dataset. 
For the GPT baselines, we still evaluated them 
by augmenting the samples corresponding to the skeleton 
with prompts in the same way as Section~\ref{secrrs}.
The results are shown in Table~\ref{tabes}. 

\subsubsection{ECE task}
Our method demonstrates the best performance 
in utterance-level causal reasoning from the 
significant improvement of our model over other baselines 
on all skeletons. Moreover, this also demonstrates that 
the conversation reasoning task not only performs poorly 
under unsupervised methods of LLMs, but also fails to 
achieve expected results under supervised methods 
tailored for relation-related tasks. 

Representing the causal relationship ``$U_{1} \rightarrow 
U_{2} \rightarrow U_{3} \rightarrow U_{4}$'', skeleton I 
serves as the foundational skeleton for the other 3 skeletons, 
with each of them adding one causal relationship to the structure 
of Skeleton I. As a result, many baselines learn the shortcuts, 
leading to better performance on Skeleton I 
compared to the other skeletons (e.g., DECN, DAG-ERC, CAE2, CAE3). 
Our approach alleviates the impact of shortcuts 
through causal strength matrix reasoning, 
resulting in better performance on Skeleton II and III. 
This corroborates the trait that Skeleton II and III, 
as the skeletons most affected by the same-speaker influence, 
(i.e., Skeleton II adds ``$U_{1} \rightarrow U_{3}$'', 
and Skeleton III adds ``$U_{2} \rightarrow U_{4}$'') 
are more easily inferred for causal relationships. 
As for Skeleton IV, it is the most difficult to identify 
because it considers additional causal relationships 
from different speakers, and it thus exhibits the worst performance 
among all skeletons in all methods. 
However, our approach exhibits less performance degradation ($4.23\% \downarrow$)
on Skeleton IV compared to other skeletons, 
relative to most other methods (GPT-3.5 $5.38\% \downarrow$, 
GPT-4 $11.25\% \downarrow$, DAG-ERC $6.36\% \downarrow$,  
CAE$_{2}$ $5.51\% \downarrow$) which indicates the superiority of 
reasoning.

\begin{figure*}
  \centering
  \subfloat[Label]{
    \includegraphics[width=0.16\linewidth]{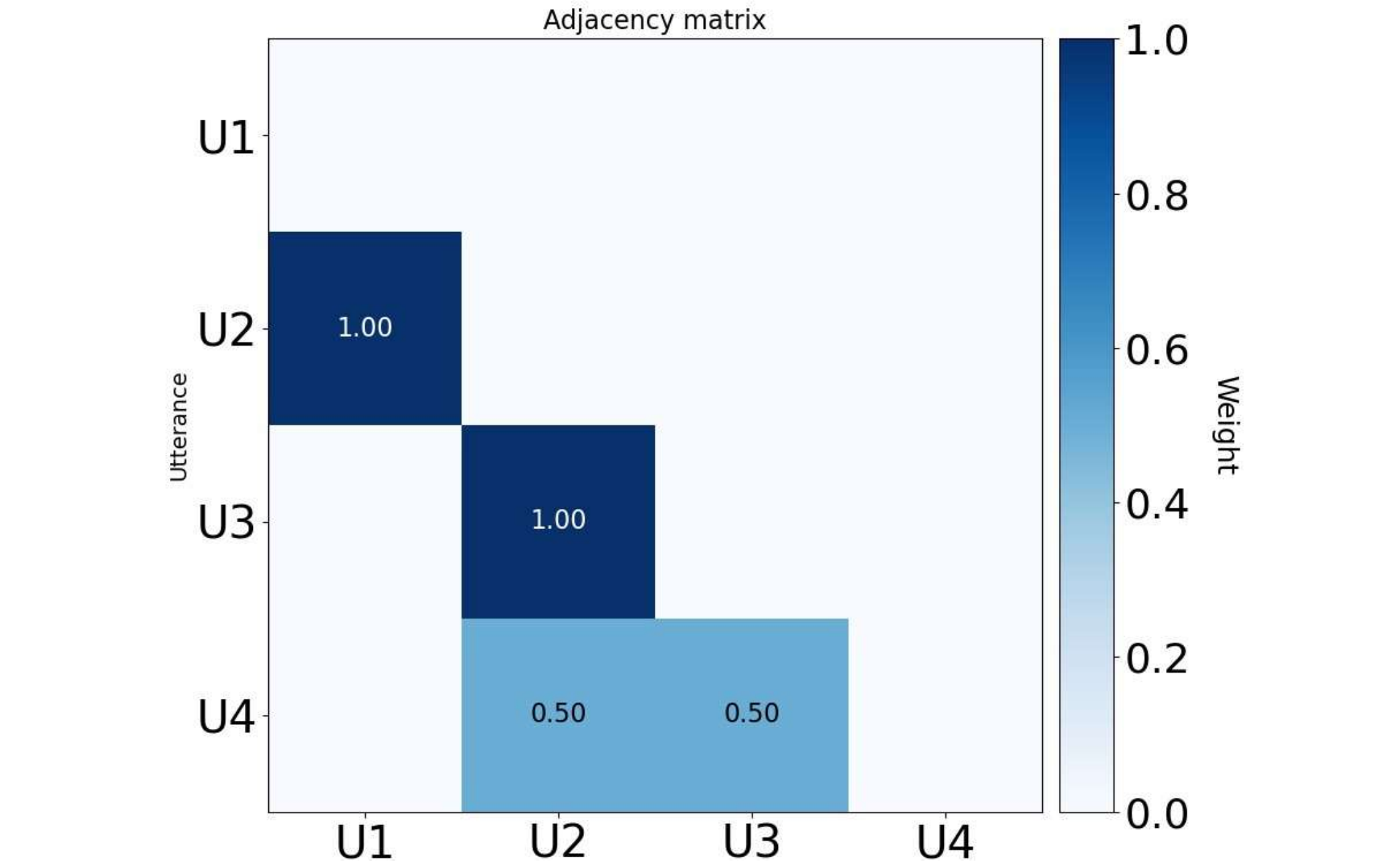}}
  \subfloat[GPT-3.5]{
    \includegraphics[width=0.16\linewidth]{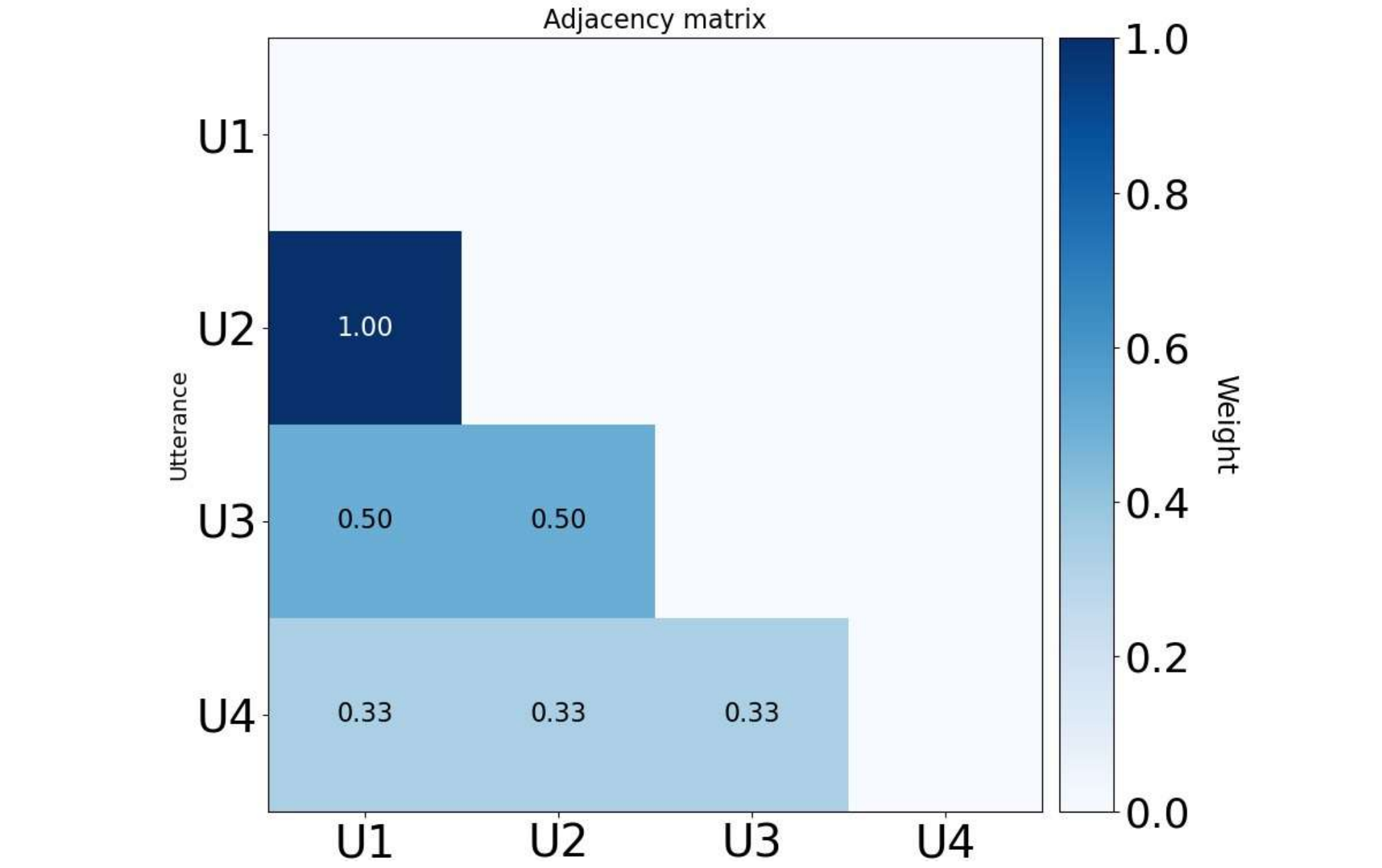}}
  \subfloat[GPT-4]{
    \includegraphics[width=0.16\linewidth]{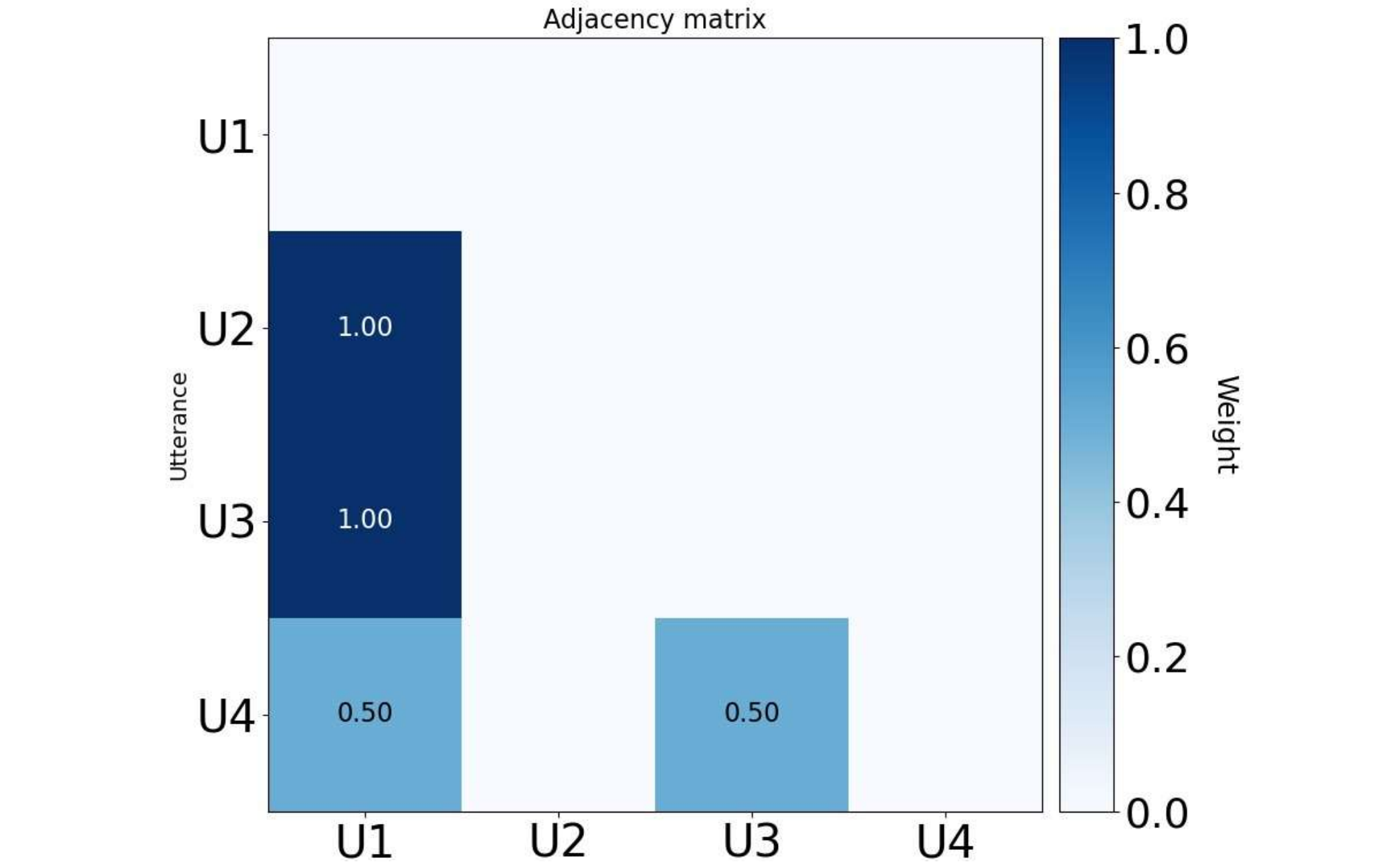}}
  \subfloat[RoBERTa]{
    \includegraphics[width=0.16\linewidth]{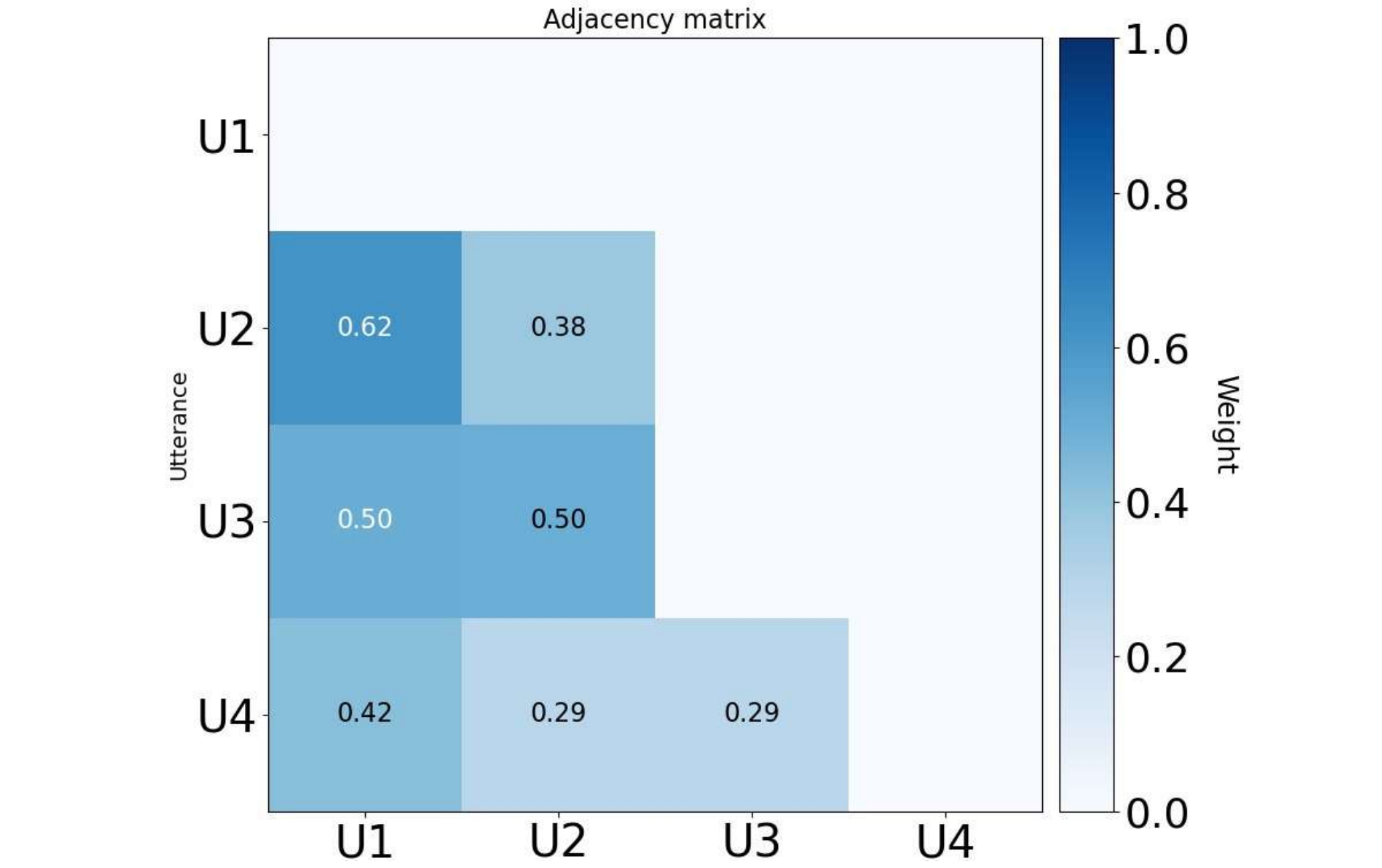}}
  \subfloat[RoBERTa$^{+}$]{
    \includegraphics[width=0.16\linewidth]{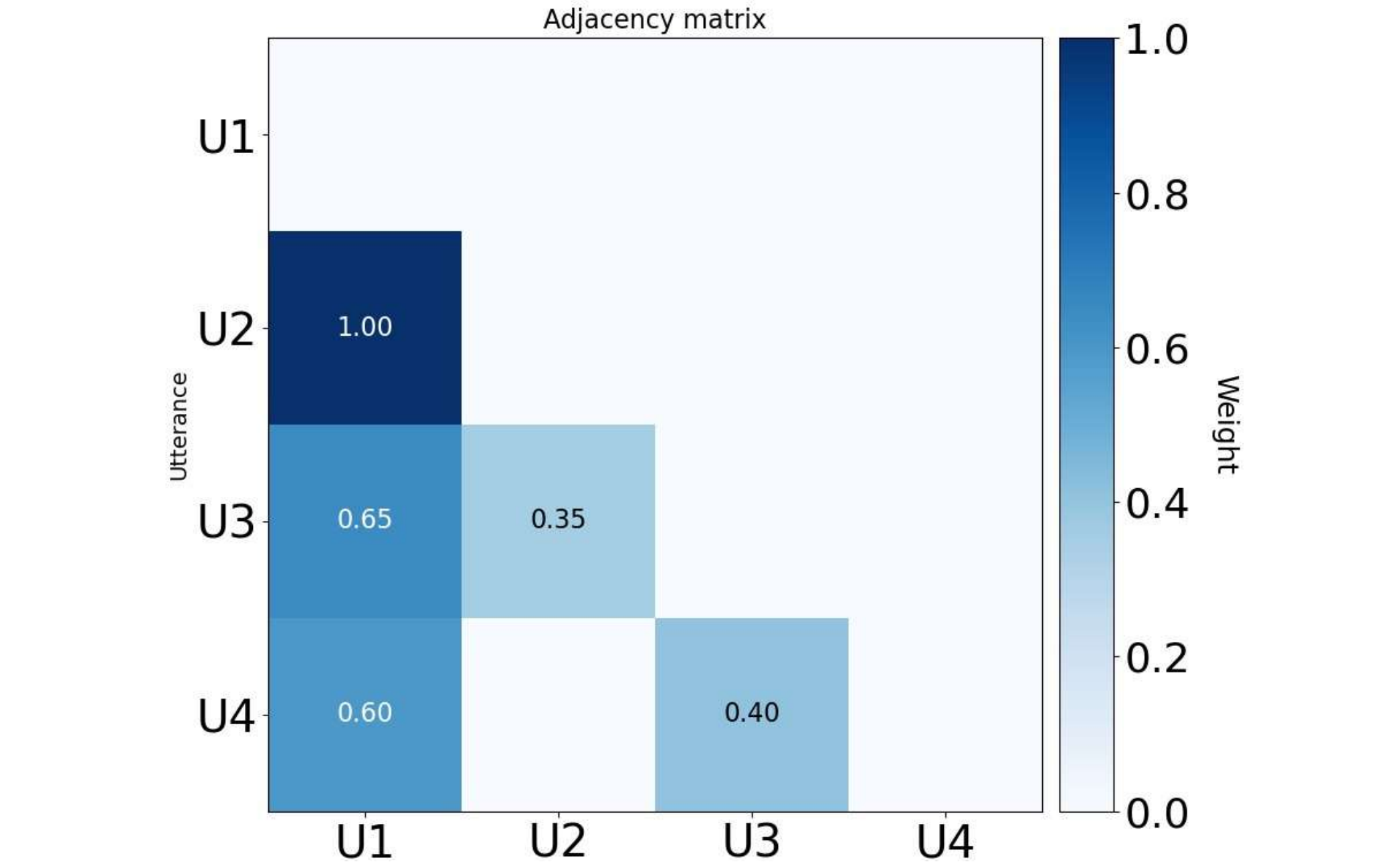}}
  \subfloat[EGAT]{
    \includegraphics[width=0.16\linewidth]{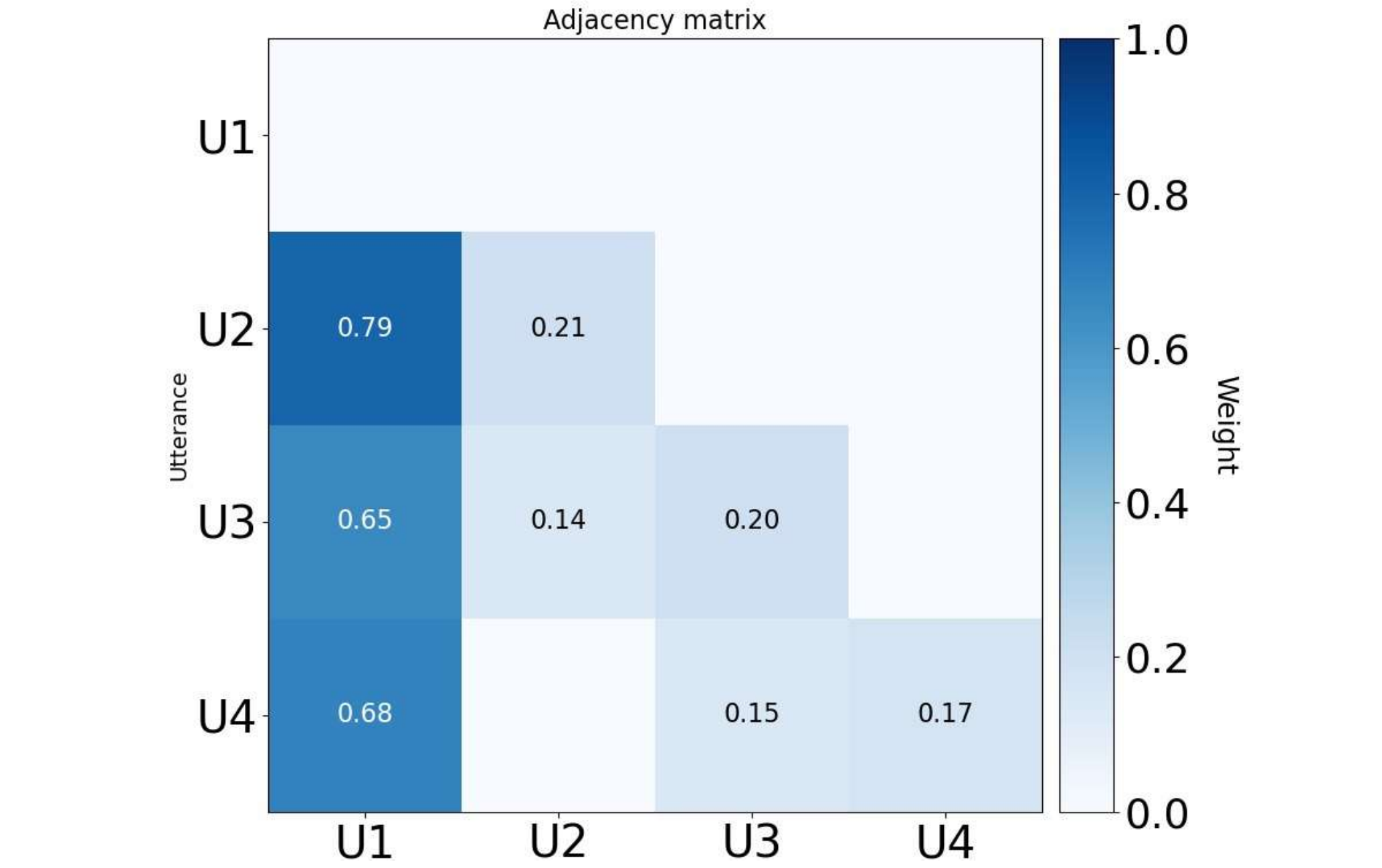}}\\
  \subfloat[DECN]{
    \includegraphics[width=0.16\linewidth]{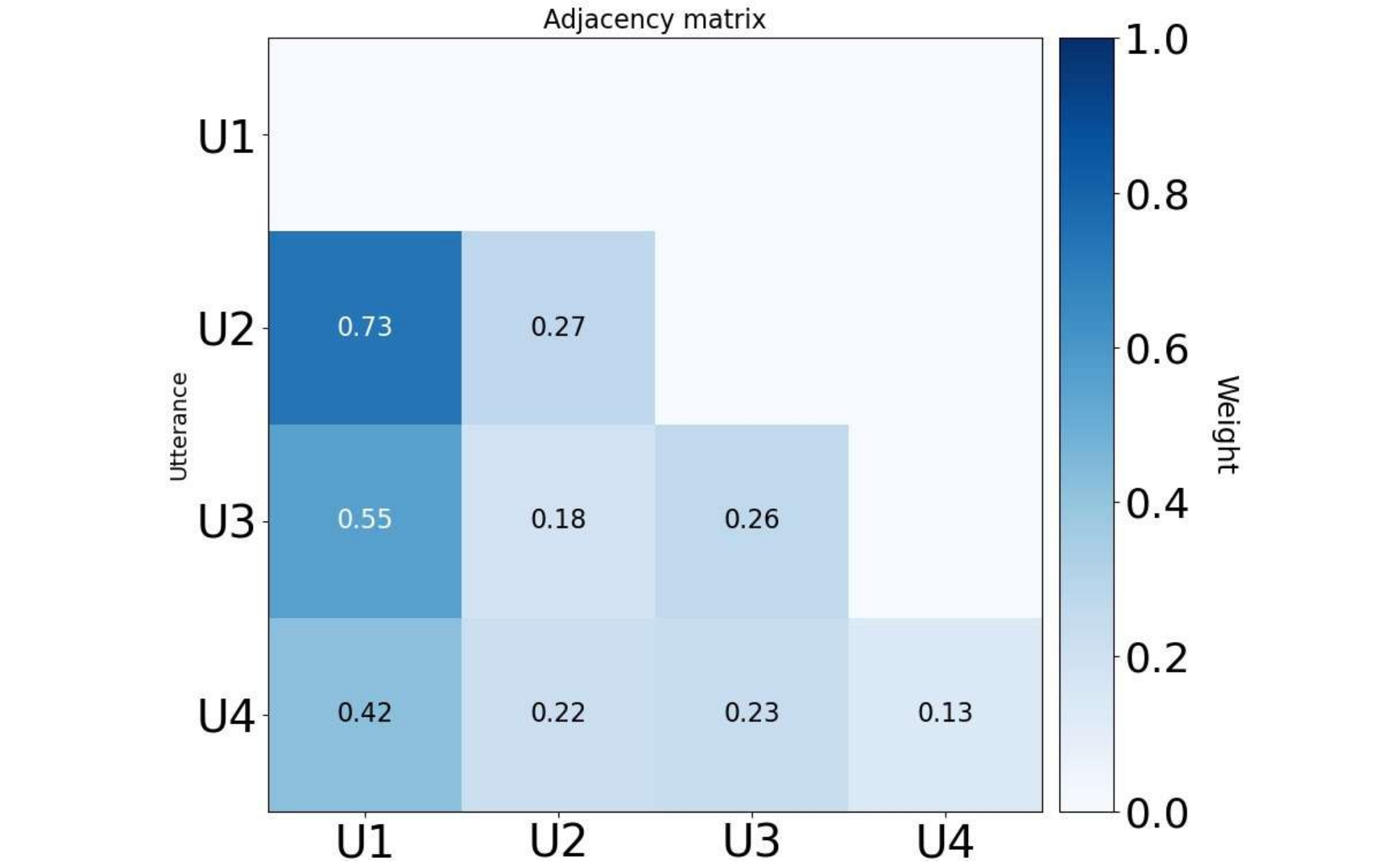}}
  \subfloat[DAG-ERC]{
    \includegraphics[width=0.16\linewidth]{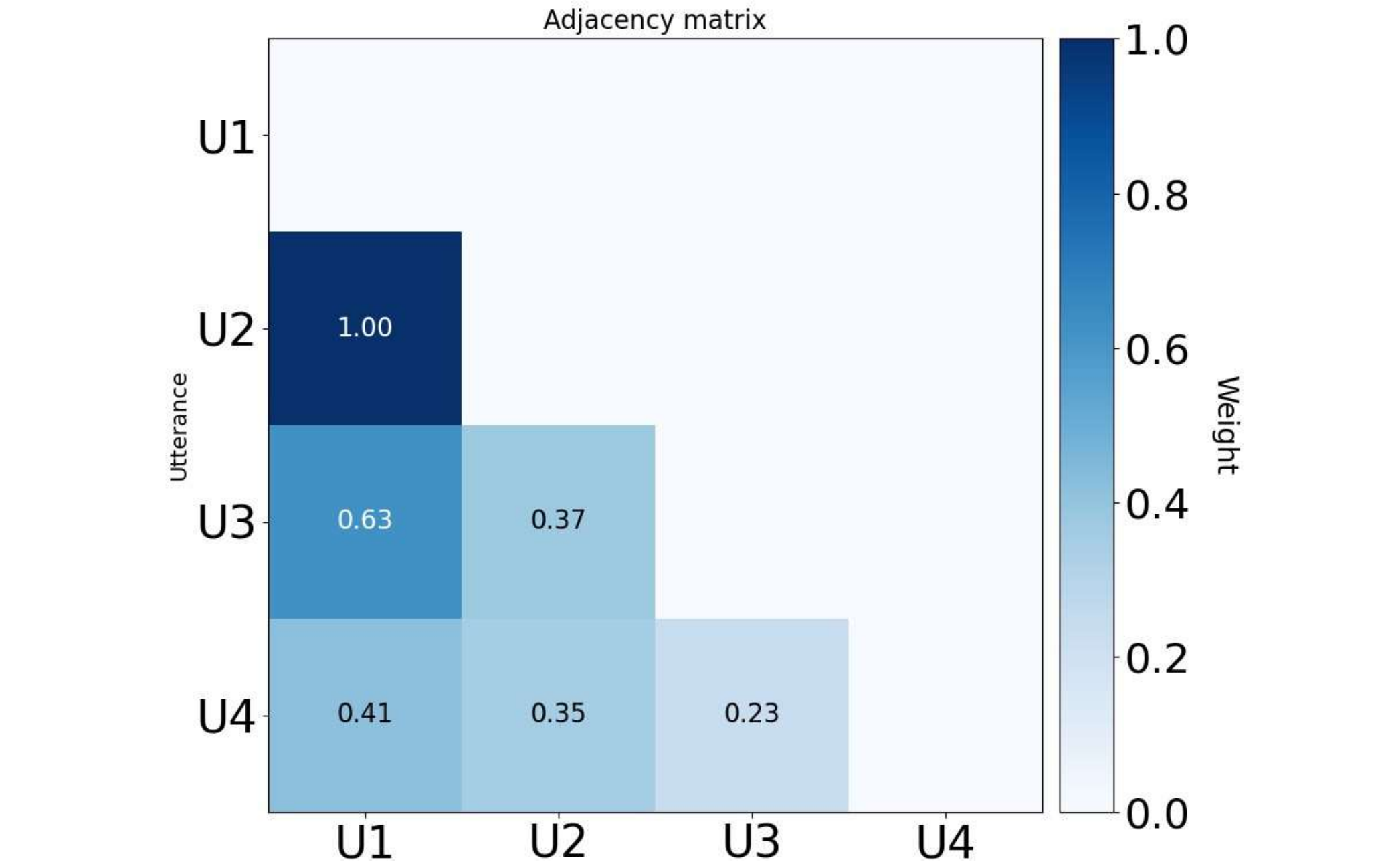}}
  \subfloat[CAE$_{1}$]{
    \includegraphics[width=0.16\linewidth]{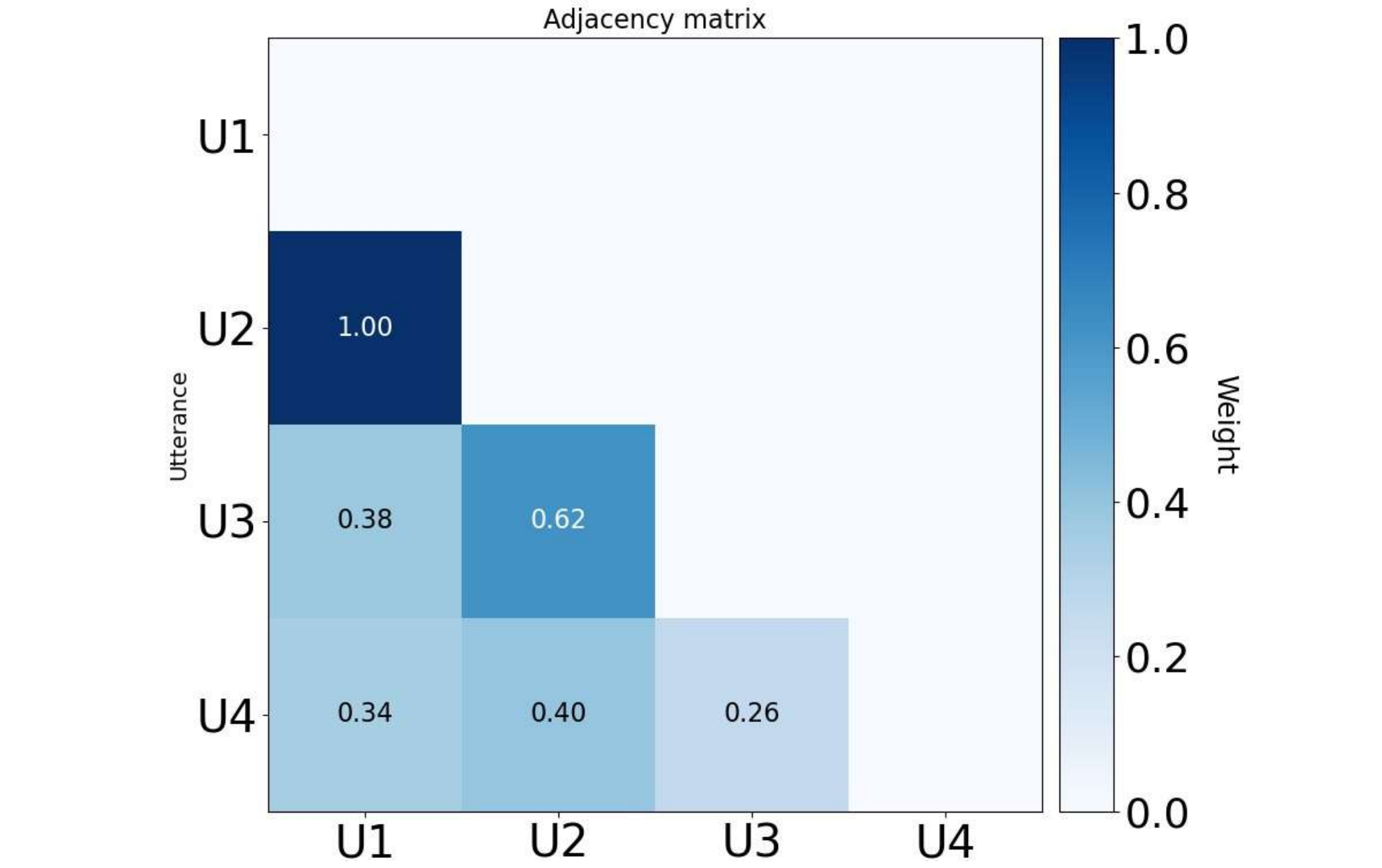}}
  \subfloat[CAE$_{2}$]{
    \includegraphics[width=0.16\linewidth]{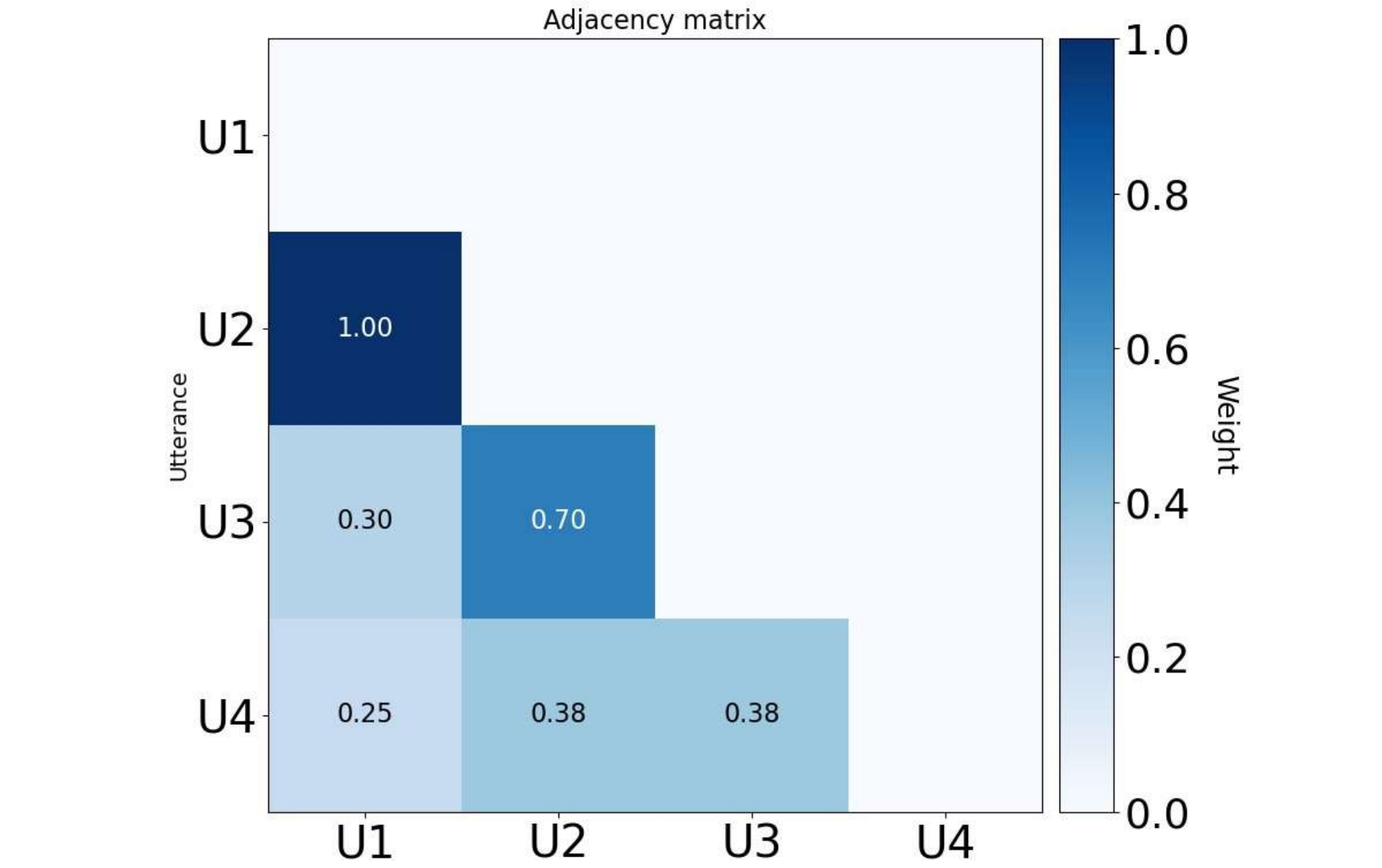}}
  \subfloat[CAE$_{3}$]{
    \includegraphics[width=0.16\linewidth]{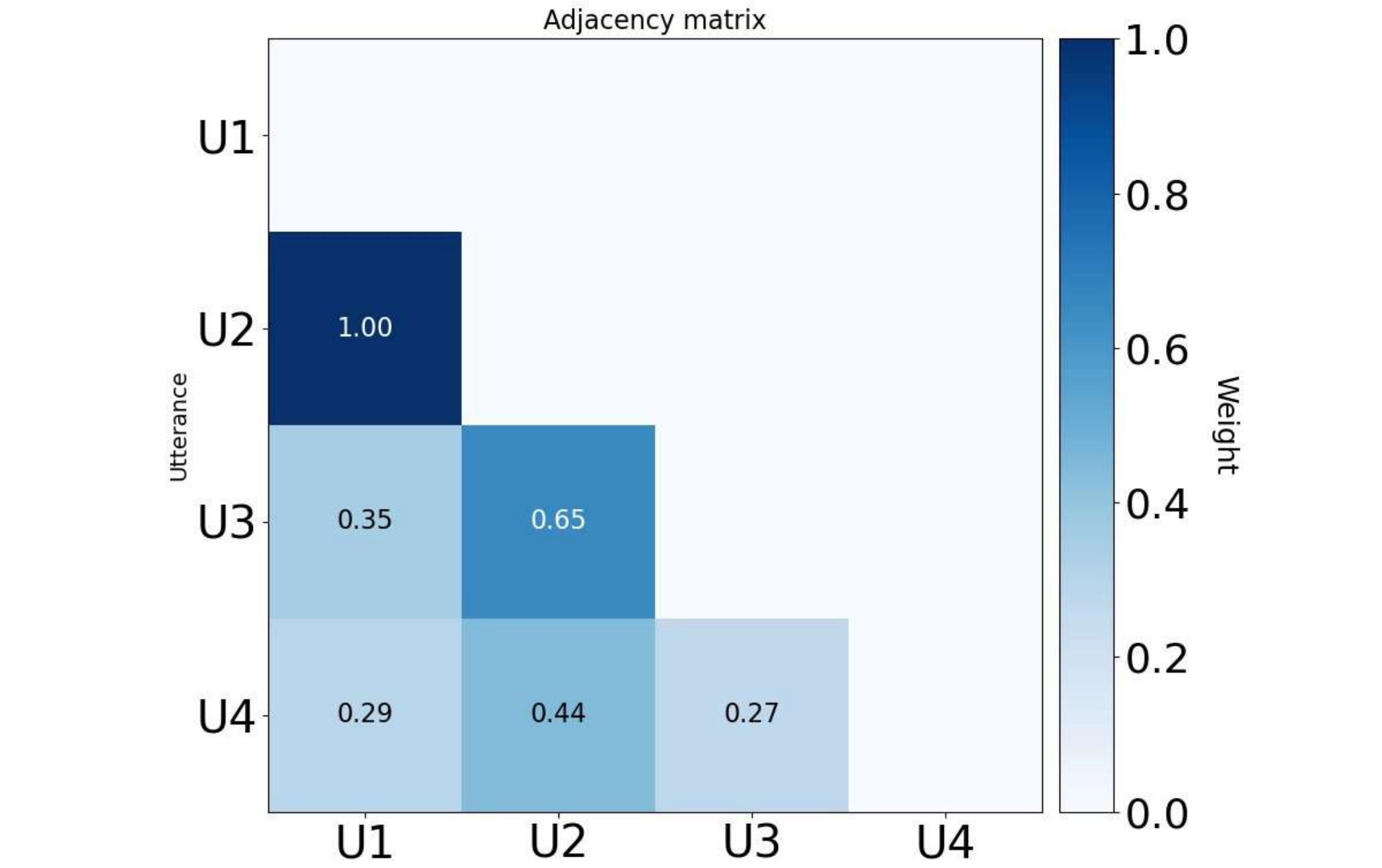}}
  \subfloat[Ours]{
    \includegraphics[width=0.16\linewidth]{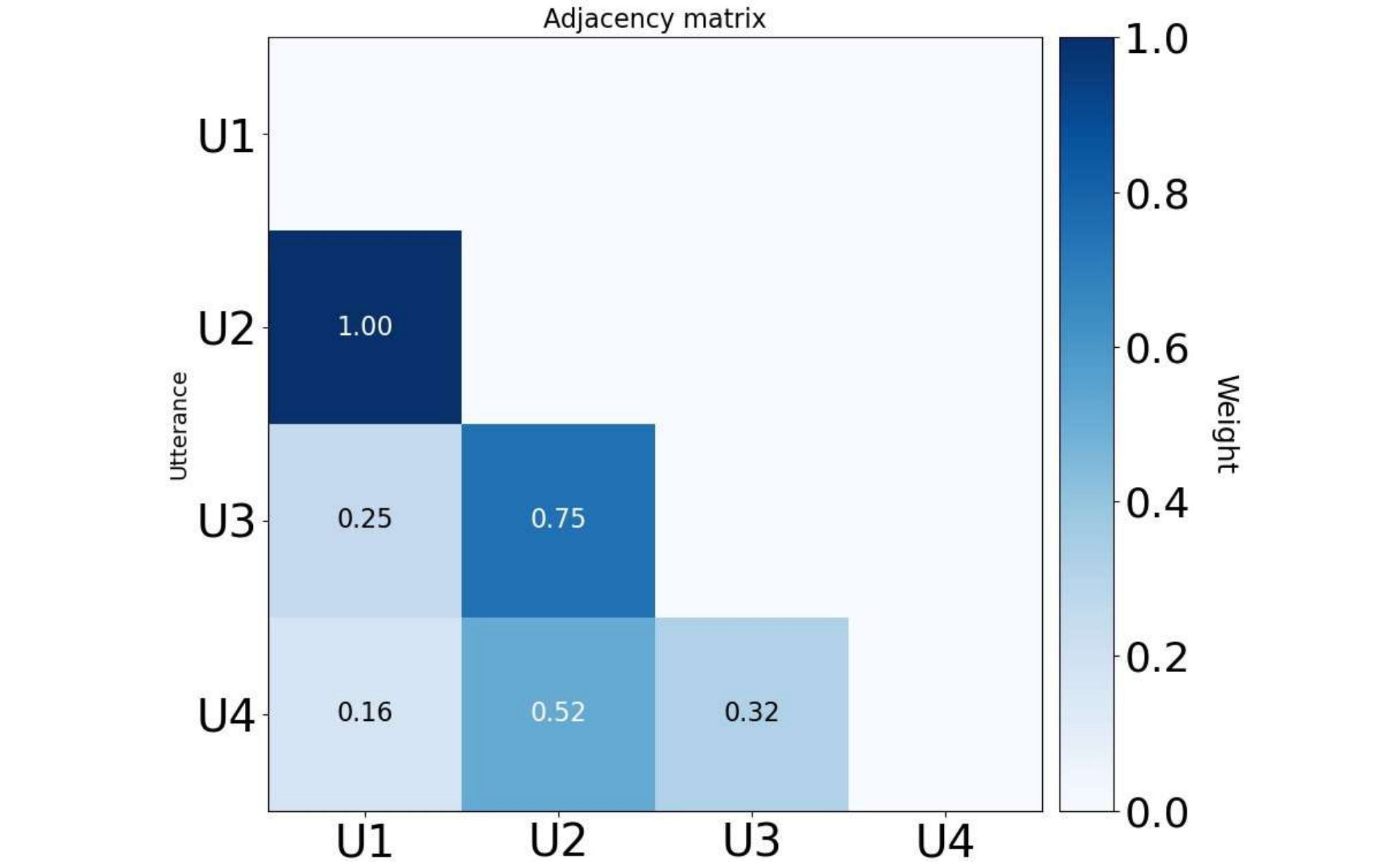}}
  \caption{Visualization of causal strength matrix $A$ of all methods. 
  Each row represents the normalized degree to which the 
  corresponding utterance is causally influenced by other utterances.}
  \label{figv}
\end{figure*}

\subsubsection{Causal Discriminability for Conversation Reasoning}

\begin{table}
  \footnotesize
  \centering
  \begin{threeparttable}   
  \resizebox{\linewidth}{!}{
    \begin{tabular}{|c|c|c|c|c|c|c|}
      \hline
      \multirow{2}{*}{Methods} & \multicolumn{2}{c}{Reversal\tnote{1}} \vline&  \multicolumn{2}{c}{Chain\tnote{2}} \vline&\multicolumn{2}{c}{Common Cause\tnote{3}} \vline\\
      \cline{2-7}
      &Pos\tnote{4}&Neg\tnote{5}&Pos&Neg&Pos&Neg\\
      \hline
       GPT-3.5&45.9 & 41.2& 43.4&44.7 & 41.6&36.4\\
       GPT-4&49.3 &46.2 &48.9 &43.7& 47.7&48.1\\
      \hline
      RoBERTa&53.9 & 52&56.4 &53.1 & 59.3&56.7\\
      RoBERTa$^{+}$&56.7 &54.9& 58.7& 56.1&52.6&54.6\\
      \hline
       EGAT& 68.7& 64.3&68.8 & 64.8&66.2&64.3 \\
       DECN&66.7&62.4&70.5&64.3&69.2&66.1\\
       DAG-ERC&71.5&68.2&72.4&64.2&69.3&66.1\\
       CAE$_{1}$&75.9&44.9&72.4&44.8&76.6&42.3\\
       CAE$_{2}$&75.8&44.1&73.4&43.9&75.1&46.2\\
       CAE$_{3}$&76.2&46.1&73.8&41.9&77.2&48.6\\
       \hline
       Ours&77.5&40.8&76.5&39.7&77.9&45.5\\
       \hline 
  
    \end{tabular}}
  \begin{tablenotes}    
    \footnotesize               
    \item[1] We set positive samples $(U_{i}, U_{j})$ 
    and negative samples $(U_{j}, U_{i})$; 
    \item[2] We set positive samples $(U_{i}, U_{k})$  and $(U_{k}, U_{j})$, 
    and negative samples $(U_{i}, U_{j})$.
    \item[3] We set positive samples $(U_{k}, U_{i})$ and $(U_{k}, U_{j})$, 
    and negative samples $(U_{i}, U_{j})$ or $(U_{j}, U_{i})$. 
  \end{tablenotes}     
\end{threeparttable} 
  \caption{Results of causal discriminability.}
  \label{tabcausaldiscri}
\end{table}

To demonstrate the superiority of our approach in conversation 
reasoning, we have evaluated the causal discriminability 
of all methods. Table~\ref{tabcausaldiscri} shows the results from 
three well-designed experiments. Taking ``Reversal'' as an example, 
the positive samples are correct causal pairs $(U_{i}, U_{j})$, 
while the negative samples are set as the corresponding 
reversed utterance pairs $(U_{j}, U_{i})$. 
For a model showcasing exemplary conversation reasoning 
capabilities, it should not only recognize the causal relationship 
of positive samples, but also distinctly differentiate 
the negative samples without causal relationship. 
Higher F1 of Pos samples imply a better extraction capability, 
while the more similar F1 scores between 
Pos and Neg indicates a weaker causal discriminability.

The results from Table~\ref{tabcausaldiscri} demonstrates that 
in all three different causal models, none of the methods (including 
LLMs like GPT-4)
were able to distinguish between negative and positive samples. 
However, our method significantly decreases the F1 score of 
negative samples indicating the effectiveness of 
incorporating implicit cause noise to enhance causal 
discriminative ability.

We illustrate this phenomenon with a practical example of results. 
For the example with cause utterance U1 and result utterance U2:

\textit{U1: ``John, your vacation must have been exciting.''}

\textit{U2: ``Yes, it was! One of the highlights of my trip was going bungee jumping.''} 

All models in Table\~ref{tabcausaldiscri} incorrectly identify 
both (1,2) and (2,1) as (result, cause) pairs. 
In a non-SCM structure, the models can only learn the semantic 
similarity between the two utterances - both U1 and U2 are 
discussing some exciting experience. Obviously, 
this information is insufficient to support more detailed 
causal recognition. However, with the SCM structure, 
our model takes into account both explicit cause $H_{U1}$ and 
implicit cause $E_{U2}$ when learning the representation $H_{U2}$. 
This means that $|H_{U1}-H_{U2}|$ will be closer to $H_{U2}$ 
and further from $H_{U1}$, thereby determining that the ``arrow'' 
points from U1 to U2. 

\subsubsection{Visualization}
To analyze how the causal analysis contributes to our approach, 
we visualized the causal strength matrix $A$ for each method 
(methods that did not calculate $A$ replaced the corresponding 
values in $A$ with their predicted results for each utterance pair). 
We masked the lower triangle of the $A$ matrix 
and normalized each row. Therefore, methods that already 
had a strict lower triangle constraint (Ours, CAE, DAG-ERC) 
have empty values on the diagonal. 

From the results, we can observe that models 
without reasoning abilities, 
whether transformer-based (GPT-3.5, GPT-4, RoBERTa, RoBERTa$^{+}$) 
or GNN-based (EGAT, DECN, DAG-ERC, CAEs), tend to construct 
relationships from the first utterance to each utterance. 
This is because from a chain causal structure perspective, 
the first utterance indirectly influences all other utterances, 
making it the most apparent shortcut in utterance relation learning.
However, our method uses SCM reasoning to enable the model 
to learn intermediate variables of causal relationships, 
thereby shifting the cause-utterance focus of the 4-th utterance 
to the correct 2-nd and 3-rd utterances.

\section{Discussion} 

We would like to further discuss the situation 
when cognitive models encounter confounding.
Confounding indicates that there is an 
unobservable variable $L$ (also called the latent variable) 
directly influencing two observable variables $X_{i}$ and $X_{j}$: 
$L \rightarrow X_{i}$ and $L \rightarrow X_{j}$. Under the confounding, 
it is intractable to determine whether the correlation 
between $X_{i}$ and $X_{j}$ is caused by the confounder $L$ 
or by their own causal relationship. 

In the creation of Simulation dataset, we assumed that the same 
speaker possesses the same \textit{system}. Specifically, $U_{1}$ 
and $U_{3}$ share the same \textit{system} and so do $U_{2}$ and 
$U_{4}$. Hence, \textit{system} should be a confounder. 

To provide a simple demonstration of the performance 
of various models on confounders, we calculated the number of 
$(U_{4}, U_{2})$ labels that were incorrectly predicted in 
Skeleton I. Non-SCM methods performed poorly: 
GPT-3 produced 242 errors, GPT-4 225 errors, RoBERTa 231 errors, 
RoBERTa$^{+}$ 233 errors, EGAT 211 errors, DECN 229 errors, 
and DAG-ERC 247 errors. However, the SCM methods 
could significantly mitigate the effects of confounders: 
CAE$_{1}$ produced 178 errors, CAE$_{2}$ 169 errors, 
CAE$_{3}$ 176 errors, while Ours only returned 155 errors. 

This mitigation can be explained through a theoretical analysis. 
In our SCM-based approach, the representation of utterance 
not only receives information from explicit causes 
(utterances in the parent set) but also from implicit causes. 
These implicit causes are treated as independent ``noise terms'' 
that are not influenced by context. 
This ensures that only the residuals of two representations 
being affected by the same implicit causes will show a correlation. 
For a linear example, if $A=E_A$, $B=fA+E_B$, 
then the residual $\Sigma_{A}=A-\lambda B=\lambda E_B \nVbar B$, 
while $\Sigma_{B}=B-\kappa A= E_B \Vbar A$. On the contrary, 
if $B=E_B$, $A=fB+E_A$, the residual $\Sigma_{A}=E_A \Vbar B$ while 
the residual $\Sigma_{B}=\lambda E_A \nVbar A$. 
However, the condition for confounding corresponds to 
$\Sigma_{A} \nVbar B$ and $\Sigma_{B} \nVbar A$. 
Therefore, our structure provides sufficient information 
for the representation to distinguish whether the correlation  
is caused by confounding or causal relationship. 

Despite the advantage our method has beyond others due 
to the variations in the probability model, 
which enhanced the i.i.d. characteristics of the noise, 
it still faces significant confounding bias.
To fully address the confounding effect, 
we further propose an experimental approach. 

With the SCM 
semantics, an intervention operation $X=x$ is represented through 
the do-operator $do(X=x)$, which performs the action of replacing 
the original distribution of $X$ by the constant distribution 
$P(X=x)=1$. That is to say, intervention aims to block all paths 
between two nodes in causal graph (called d-separate). 
Using intervention, we can analyze 2 confounding situations: 
confounding between non-adjacent nodes and between adjacent nodes. 

\textbf{Confounding between Non-adjacent Nodes}: 
Consider two utterances $U_{i}$ and $U_{j}$ being non-adjacent nodes. 
Let $Pa$ be the union of the parents of $U_{i}$ and $U_{j}$: 
$Pa=U_{i} \cup U_{j}$. If we perform an intervention on $Pa$ 
(i.e., $do(Pa=pa)$), we thus have $U_{i} \nVbar U_{j}$ if and only 
if there is a latent confounder $L$ such that 
$U_{i} \leftarrow L \rightarrow U_{j}$. 

\textbf{Confounding between Adjacent Nodes}: 
Consider two utterances $U_{i}$ and $U_{j}$ being adjacent nodes: 
$U_{i} \rightarrow U_{j}$. If there are no latent confounders, we 
have $P(U_{j}|U_{i})=P(U_{j}|do(U_{i}=u_{i}))$. 

However, 
this is only a theoretical design proposed through causal analysis, 
as it is difficult to implement in this paper 
for the following reasons: a) in our dataset, there are 
no non-adjacent utterances as all utterances are constructed 
based on a chain structure. This makes it impossible for us 
to test for confounding between non-adjacent nodes; b) to achieve 
$P(X=x)=1$ in intervention, we must calculate the conditional 
probabilities between each utterance by adjusting Z formula
~\cite{pearl2009causality}, 
treating each utterance as an ancestor sample from the 
Bayesian network. Although we simulated this process 
via variational inference, computing the probability of 
high-dimensional vectors remains infeasible to infer. 
All in all, we will attempt to address the issue of 
confounding factors in future work. 

\section{Conclusion}
In this paper, we aim to address the challenge of 
conversation reasoning, which has proven to be a difficult task 
for both supervised methods and unsupervised methods (LLMs). 
Firstly, we construct a Conversation Cognitive Model (CCM) 
based on intuitive theories, which incorporate concepts 
related to dialogue such as utterances, plans, actions, 
and mental states. This model reflects the factors 
that give rise to utterances and the effects that utterances have, 
which fills the gap in the field of conversational reasoning 
by providing a complete and grounded cognitive model 
that was previously missing. 
Secondly, we algebraically transform this CCM into 
a Structural Causal Model (SCM), which represents the 
complex relationships between utterances through explicit 
and implicit relationships in a graphical structure. 
This effect makes it possible to perform statistical modeling 
and causal analysis of various concept-level factors in conversations.
Thirdly, we instantiate the SCM using variational inference, 
treating the implicit causes as latent variables, 
to generate a more favorable causal representation of utterances 
and an inferable causal graph. Finally, 
we release a synthetic dataset with implicit causes 
and a simulated dataset with complete causal relationships, 
filling the last remaining gap in the field of 
conversation reasoning. We conducted extensive experiments 
to internally and externally evaluate our model, 
and discussed the issue of latent confounders 
that cannot be resolved at present, 
proposing some experimental designs as a starting point 
for other researchers to gain insights. 
\bibliographystyle{IEEEtran}
\bibliography{custom}

\end{document}